\documentclass[10pt,twocolumn,letterpaper]{article}

\usepackage{cvpr}              

\usepackage[dvipsnames]{xcolor}
\newcommand{\red}[1]{{\color{red}#1}}
\newcommand{\blue}[1]{{\color{blue}#1}}

\definecolor{howard_green}{rgb}{0.0, 0.5, 0.0}

\usepackage{graphicx}
\usepackage{amsmath}
\usepackage{amssymb}
\usepackage{booktabs}
\usepackage{tabularx}
\usepackage{multirow}
\usepackage{xcolor}
\usepackage{pifont}
\usepackage[normalem]{ulem}

\newcommand{\cmark}{\ding{51}}%
\newcommand{\xmark}{\ding{55}}%

\definecolor{cvprblue}{rgb}{0.21,0.49,0.74}
\usepackage[pagebackref,breaklinks,colorlinks,citecolor=cvprblue]{hyperref}

\usepackage[capitalize]{cleveref}
\crefname{section}{Sec.}{Secs.}
\Crefname{section}{Section}{Sections}
\Crefname{table}{Table}{Tables}
\crefname{table}{Tab.}{Tabs.}

\def\paperID{6385} 
\def\confName{CVPR}
\def\confYear{2024}

\title{Boosting Flow-based Generative Super-Resolution Models via Learned Prior}

\author{Li-Yuan Tsao$^{1}$ Yi-Chen Lo$^{2}$ Chia-Che Chang$^{2}$ Hao-Wei Chen$^{1}$ Roy Tseng$^{2}$ Chien Feng$^{1}$ Chun-Yi Lee$^{1}$\\
$^{1}$National Tsing Hua University\hspace{5pt} $^{2}$MediaTek Inc.\\
{\tt\small \{lytsao, jaroslaw1007, kmes9961002\}@gapp.nthu.edu.tw} \\
{\tt\small{\{{yichen.lo, chia-che.chang, roy.tseng}\}@mediatek.com\hspace{5pt}
cylee@cs.nthu.edu.tw}}
}

\begin{document}
\maketitle

\begin{abstract}
Flow-based super-resolution (SR) models have demonstrated astonishing capabilities in generating high-quality images. However, these methods encounter several challenges during image generation, such as grid artifacts, exploding inverses, and suboptimal results due to a fixed sampling temperature. To overcome these issues, this work introduces a conditional learned prior to the inference phase of a flow-based SR model. This prior is a latent code predicted by our proposed latent module conditioned on the low-resolution image, which is then transformed by the flow model into an SR image. Our framework is designed to seamlessly integrate with any contemporary flow-based SR model without modifying its architecture or pre-trained weights.
We evaluate the effectiveness of our proposed framework through extensive experiments and ablation analyses. The proposed framework successfully addresses all the inherent issues in flow-based SR models and enhances their performance in various SR scenarios. Our code is available at: \url{https://github.com/liyuantsao/BFSR}
\vspace{-10pt}
\end{abstract}

\section{Introduction}
\label{sec::introduction}
\vspace{-3pt}
Image super-resolution (SR) aims to reconstruct a high-resolution (HR) image given its low-resolution (LR) counterpart. Typically, the effectiveness of SR methods is evaluated based on fidelity and perceptual quality of the generated SR images, with commonly used metrics such as Peak Signal-to-Noise Ratio (PSNR), Structural Similarity (SSIM) for the former, and Learned Perceptual Image Patch Similarity (LPIPS)~\cite{lpips} for the latter. However, optimizing both metrics simultaneously is fundamentally challenging due to the inherent perception-distortion trade-off~\cite{tradeoff} in SR tasks. As a result, existing SR methods are broadly classified into two categories: \textit{fidelity-oriented SR}, which prioritizes pixel-wise reconstruction accuracy, and \textit{generative SR}, which focuses on enhanced visual quality. 

The recent emergence of Flow-based SR~\cite{srflow, hcflow, adflow, ncsr, fsncsr, linf} bridges this divide, as flow models possess the capability to control the diversity of image content during inference time by adjusting the sampling temperature (\ie., the standard deviation of a Gaussian distribution). As a result, a single flow-based SR model (or simply ``\textit{flow model}’’ hereafter) can produce images that either prioritize high fidelity or exhibit improved perceptual quality. This unique feature provides flow-based models with the potential to excel in each category of SR methods, making them a promising framework for SR tasks.
\begin{figure}[t]
  \centering
  \includegraphics[width=1\linewidth]{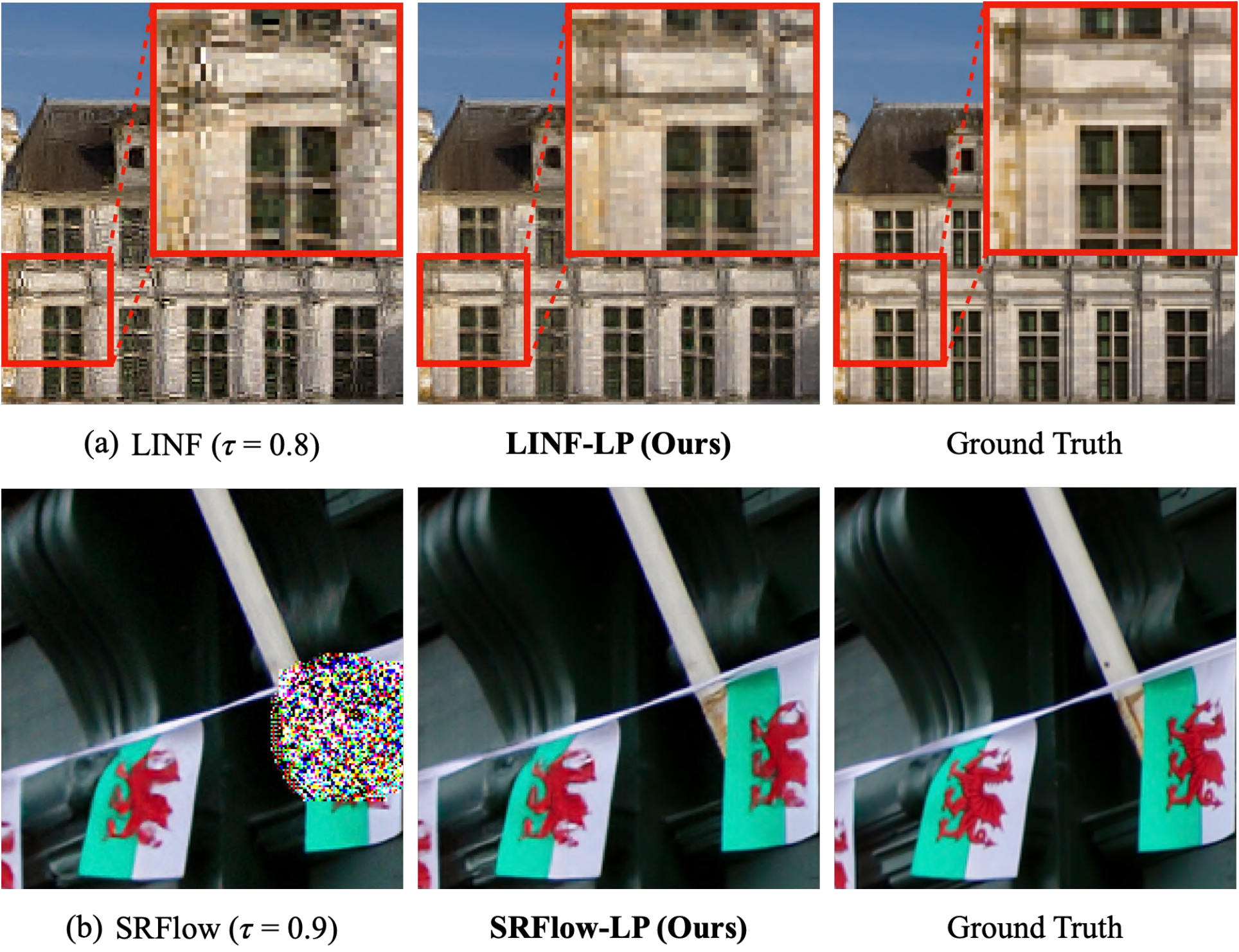}
  \caption{
    Our framework enhances the capability of contemporary flow-based SR models by: (1) mitigating grid artifacts, (2) preventing exploding inverses, and (3) eliminating the use of a fixed sampling temperature to unlock the full potential of flow models.
  }
  \label{fig:teaser}
  \vspace{-16pt}
\end{figure}


Despite their flexibility, flow-based SR methods encounter several challenges in the image generation process. These include (1) grid artifacts in the generated images, (2) the exploding inverse issue, and (3) suboptimal results stemming from the use of a fixed sampling temperature.
``Grid artifacts’’ stands for the discontinuities in textures within an image~\cite{details}. As depicted in the left image of Fig.~\ref{fig:teaser}~(a), distinct borders between the generated image patches could occasionally lead to a checkerboard pattern~\cite{deconv}. 
Another critical aspect is the ``exploding inverse"~\cite{explode, robust} in invertible neural networks, which refers to the occurrence of infinite values in the inverse process. This phenomenon leads to the appearance of noisy patches within images. As illustrated in Fig.~\ref{fig:teaser}~(b), images affected by exploding inverse display some noisy patches, which obscure the original content within the image.
\begin{figure}[t]
  \centering
  \includegraphics[width=1\linewidth]{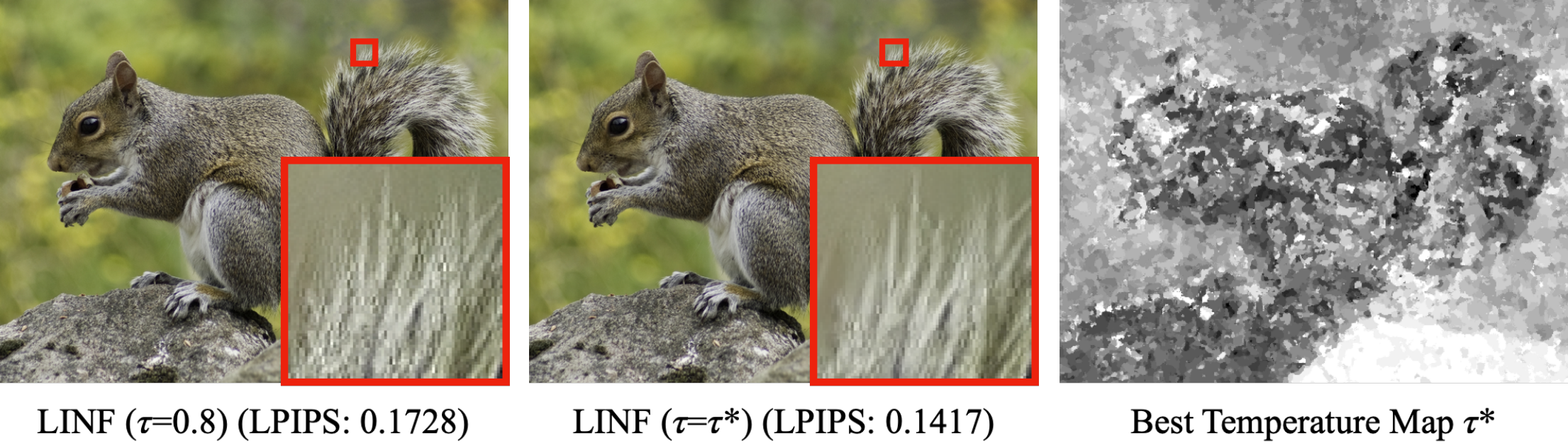}
  \caption{
    The values in the ``best temperature map” represent the sampling temperature for each position that yields the optimal LPIPS score, where a lighter color represents a higher temperature. The best temperatures are determined through an exhaustive search, with the LPIPS scores calculated on the whole image.
  }
  \vspace{-14pt}
  \label{fig:temp_map}
\end{figure}
Moreover, despite using a fixed sampling temperature during evaluation is a common practice in flow-based SR methods, this approach might not always be suitable, as the ideal temperature settings could vary across different regions.  As depicted in Fig.~\ref{fig:temp_map}, the areas of low-frequency components such as backgrounds favor higher temperatures for diversity enhancement. In contrast, high-frequency areas (\eg, the detailed fur of a squirrel) require lower temperatures to maintain consistent contents. While the optimal temperature for each area can be identified through an exhaustive search to boost performance, this approach may be highly cost-ineffective and impractical in real-world scenarios. As a result, the efficacy of flow models could be constrained by the suboptimal temperature setting. 
Based on these challenges, flow-based SR methods still hold the potential for further enhancement.

\begin{figure}[t]
  \centering
  \includegraphics[width=1\linewidth]{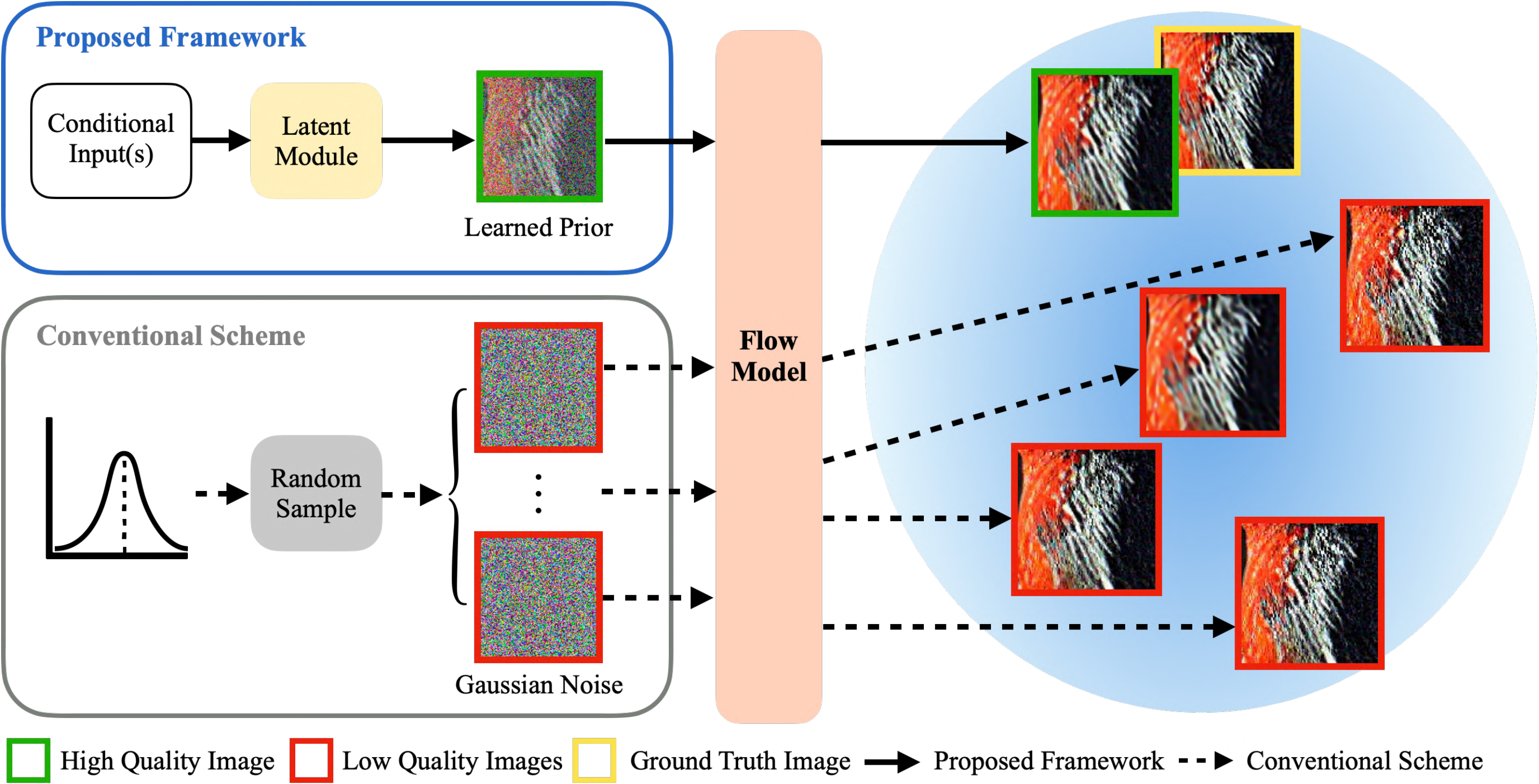}
  \caption{
    The proposed framework leverages a conditional learned prior to the inference phase of a flow-based SR model, aiming to approximate the ideal latent code corresponding to the HR image.
  }
  \label{fig:framework_concept}
  \vspace{-12pt}
\end{figure}

In light of the aforementioned issues, a key element to addressing these problems of flow models could be a learned prior. This learned prior should hold the following properties. First, it captures the correlation between image patches to alleviate the discontinuities. Second, this prior should be prevented from being an out-of-distribution input to the flow model, which leads to an exploding inverse~\cite{explode, robust}. Third, to eliminate the need to fine-tune a sampling temperature, this prior should be directly generated by a model instead of sampling from a Gaussian prior.
Fig.~\ref{fig:framework_concept} illustrates the integration of a learned prior into SR tasks. In this approach, the flow model receives the learned prior as input, replacing the conventional method which employs a randomly sampled latent code.

To achieve this, this study proposes a framework that introduces a latent module as the conditional learned prior. The latent module is designed to directly estimate a learned latent code in a single-pass for flow model inference. This design philosophy aligns well with the principle that real-world SR applications prefer fast inference and consistent predictions~\cite{fsrcnn, carn}. Specifically, the proposed framework consists of two main components: a latent module and a flow model, with the latter being any contemporary flow-based SR model. The latent module is responsible for predicting a latent code conditioned on the LR image, which is then transformed by the flow model into an SR image.

In this work, the proposed latent module is integrated with two flow-based SR models, including the arbitrary-scale SR framework LINF~\cite{linf} and  the fixed-scale framework SRFlow~\cite{srflow}, to demonstrate its effectiveness, generalizability, and flexibility with extensive experiments. 
%
The contribution of this study can be summarized as follows:
\begin{itemize}
\item This study reveals three inherent deficiencies in flow-based SR methods, which are key issues that require enhancements to fully unleash the potential of flow models.
\item We introduce a conditional learned prior to the inference phase of a flow-based SR model, which effectively addresses the inherent issues of flow models without modifying their architecture or pre-trained weights.
\item The proposed latent module is highly flexible in terms of network architecture design, which allows the adoption of any commonly used backbone. Even a lightweight module could introduce noticeable improvements.
\item Our proposed framework generalizes to both fixed-scale and arbitrary-scale flow-based SR frameworks without requiring customized components, which also leads to advancements at out-of-training-distribution scales.
\end{itemize}

\vspace{-3pt}
\section{Related Work}
\label{sec::related_work}
\vspace{-3pt}
\subsection{Image Super-Resolution}
\vspace{-3pt}
Contemporary deep learning-based SR methods~\cite{srcnn, espcn, vdsr, lapsrn, rcan, dbpn} primarily fall into two categories: fidelity-oriented SR and generative SR.
Fidelity-oriented SR methods~\cite{edsr, rdn, swinir, hat, omnisr, srformer, craft} aim to promote pixel-wise reconstruction accuracy and are commonly trained with L1 or L2 loss. Despite achieving high fidelity, these methods often produce blurry results. 
To overcome this and synthesize realistic HR images, various generative SR techniques have been proposed, including GAN-based~\cite{srgan, sftgan, esrgan, ranksrgan, fxsr, srooe}, AR-based~\cite{larsr}, flow-based~\cite{srflow, hcflow, adflow, ncsr, fsncsr}, and diffusion-based~\cite{sr3, srdiff} methods. Although these works effectively generate diverse and visually plausible HR images, they are limited to super-resolving images at a predefined upsampling scale, restricting their real-world applicability. Recently, various arbitrary-scale SR methods~\cite{metasr, liif, lte, clit, ciaosr, itsrn, ultrasr, linf} have emerged. These approaches are capable of producing high-fidelity images even at significantly large upsampling scales (\eg, 30$\times$). Among these methods, LINF~\cite{linf} pioneers the field of arbitrary-scale generative SR, which is able to generate realistic images across continuous scales.

\subsection{Flow-based Super-Resolution}
\vspace{-3pt}
Flow-based SR methods utilize normalizing flows to establish a bijective mapping between the conditional distribution of HR images and a prior distribution. This approach addresses the ill-posed nature of SR tasks by modeling the entire HR image space. Besides excelling in general SR tasks~\cite{srflow, hcflow, adflow, ncsr, fsncsr, linf}, flow-based SR methods find applications in a variety of specialized SR tasks. These include blind image SR~\cite{fkp}, remote sensing image~\cite{blindsrsnf}, Magnetic Resonance Imaging~\cite{mriflow}, Magnetic Resonance Spectroscopic Imaging~\cite{mrsiflow}, and scientific data SR~\cite{psrflow}.

\subsection{Learned Prior}
\vspace{-3pt}
Modern deep generative models~\cite{bond2021deep} typically learn mappings between data and latent variables (i.e., latent codes) based on Gaussian prior, either explicitly such as variational autoencoders (VAEs)~\cite{vae}, normalizing flows~\cite{kobyzev2020normalizing}, diffusion models~\cite{sohl2015deep, ho2020denoising, song2019generative, song2020improved, song2020score} or implicitly such as generative adversarial networks (GANs)~\cite{gan, karras2019style}. Regardless of their distinct learning formulations, they generally adopt a fixed Gaussian prior for sampling during inference. Recently, improved VAEs~\cite{rezende2015variational, kingma2016improved, dilokthanakul2016deep, chen2016variational, gulrajani2016pixelvae, van2017neural, tomczak2018vae, razavi2019generating, ghosh2020from, esser2021taming, chadebec2022pythae} have highlighted more expressive priors for better lower bound and sample generation. These advancements suggest that learning priors using neural networks can be used for sampling since the decoder network was effectively trained on latent codes from the learned posterior. 
Likewise, we propose to learn a latent module as a form of learned prior over latent codes inverted from training data by conditional normalizing flow, to remedy the train-test gap as a general framework to boost flow-based super-resolution quality.


\vspace{-3pt}
\section{Preliminaries}
\label{sec::preliminaries}
\vspace{-3pt}
\subsection{Fixed-scale Flow-based SR}
\vspace{-3pt}
Given an LR image $x \in \mathbb{R}^{H \times W \times 3}$ and an HR image $y \in \mathbb{R}^{sH \times sW \times 3}$ with a predefined scaling factor $s$, fixed-scale flow-based SR models \cite{srflow, hcflow, adflow, ncsr} aim to capture the entire conditional distribution $p_{Y|X}(y|x)$. Specifically, they learn a bijective mapping between a conditional distribution $p_{Y|X}(y|x)$ and a prior distribution $p_Z(z)$, where $z \in \mathbb{R}^{sH \times sW \times 3}$ is typically 
a standard normal distribution $\mathcal{N}(0, I)$. By utilizing an invertible neural network with $k$ invertible layers, such transformation is given by:
\vspace{-5pt}
\begin{equation}
\begin{split}
    z &= f_{\theta}(y; x) = f_{k} \circ \cdot\cdot\cdot \circ f_{1}(y; x), \\
    y &= f^{-1}_{\theta}(z; x) = f^{-1}_{1} \circ \cdot\cdot\cdot \circ f^{-1}_{k}(z; x). \\
\end{split}
\label{eq:transformation}
\end{equation}
Additionally, according to the change of variable theorem, the probability of an HR-LR image pair ($y$, $x$) is defined as: \vspace{-10pt}
\begin{equation}
\begin{split}
p_{Y|X}(y|x, \theta) &= p_{Z}(f_{\theta}(y;x)) \cdot \Big| \det \frac{\partial f_{\theta}(y;x)}{\partial y}\Big|.
\end{split}
\label{eq:logp_fixed}
\vspace{-5pt}
\end{equation}
During training, fixed-scale flow-based SR models can be optimized through negative log likelihood (NLL) loss with a large set of HR-LR training pairs $ \{(y_i, x_i)\}^{N}_{i=1}$.

\subsection{Arbitrary-scale Flow-based SR}
\vspace{-3pt}
Despite successfully tackling the ill-posed nature of SR task by modeling the HR image space, fixed-scale flow-based SR models are only able to super-resolve images with a predefined scaling factor (\eg, 4$\times$), limiting their practicality.
To address this issue, the arbitrary-scale flow-based SR framework ``\textit{Local Implicit Normalizing Flow}'' (LINF)~\cite{linf} shifts the learning target from the entire HR image to 
local patches. 
During inference, it generates local patches at corresponding coordinates 
independently, and then combines these patches to form the final image. 

Specifically, given an LR image $x \in \mathbb{R}^{H \times W \times 3}$, the coordinate of a local patch $c_{i,j} \in \mathbb{R}^2$, a scaling factor $s$, and the corresponding HR patch $y_{i,j} \in \mathbb{R}^{n \times n \times 3}$, where $n$ is typically set to 3, and $i, j$ is the index of an image patch. The goal of LINF \cite{linf} is to learn a bijective mapping between a conditional distribution $p_{Y|X}(y_{i,j}|x, c_{i,j}, s)$ and a latent distribution $p_{Z}(z) = \mathcal{N}(0, I)$, where $z \in \mathbb{R}^{n \times n \times 3}$. Similar to fixed-scale framework, the probability is given by:
\vspace{-5pt}
\begin{equation}
\begin{split}
p_{Y|X}(y_{i,j}|x, c_{i,j}, s, \theta) =  \quad\quad\quad\\
p_{Z}(f_{\theta}(y_{i,j};x, c_{i,j}, s)) \cdot \Big| \det &\frac{\partial f_{\theta}(y_{i,j};x, c_{i,j}, s)}{\partial y_{i,j}}\Big|.
\end{split}
\label{eq:logp_arbitrary}
\vspace{-10pt}
\end{equation}

In practice, the HR patch $y_{i,j}$ is replaced by the residual map $m_{i,j} = y_{i,j} - x_{i,j}^{\textup{up}}$, where $x_{i,j}^{\textup{up}}$ represents the bilinear-upsampled LR image. During training, LINF \cite{linf} utilizes the Negative Log-Likelihood (NLL) loss along with pixel-wise L1 loss and VGG perceptual loss~\cite{vggloss} for additional fine-tuning across various metrics.



\vspace{-3pt}
\section{Methodology}
\label{sec::methodology}
\vspace{-3pt}
In this section, we begin by defining the formulation of the proposed method, followed by a detailed description of the proposed framework and an elaboration on the design of the objective function.
\subsection{Problem Formulation}
\vspace{-3pt}
Given an LR image $x \in \mathbb{R}^{H \times W \times 3}$ and a flow model $f_\theta$,  the objective of this work is to derive a latent code $z^* \in \mathbb{R}^{sH \times sW \times 3}$, given by: $y = f_\theta^{-1}(z^*; x),$
where $y \in \mathbb{R}^{sH \times sW \times 3}$ represents the ground truth HR image, which is not available during inference, and $s$ represents a scaling factor. 
Successfully identifying $z^*$ enables the precise reconstruction of the corresponding HR image $y$. This process is facilitated by the invertible nature of normalizing flow, which guarantees the existence of $z^*$. To realize this objective, we utilize a latent module $G$ designed to generate a latent code $\hat{z}$ in a single-pass during inference, which aims to approximate $z^*$ as closely as possible, expressed as: 
\vspace{-5pt}
\begin{equation}
    z^* \approx \hat{z} = G(\cdot).
\label{eq:fixed_objective}
\vspace{-5pt}
\end{equation}

%
Similarly, for the arbitrary-scale SR framework, we aim to identify latent codes $\hat{z}_{i, j}$ such that $\hat{z}_{i, j} = z^*_{i, j}$ for all $i, j$ within an image, formulated as: 
\vspace{-5pt}
\begin{equation}
    z^*_{i, j} \approx \hat{z}_{i, j} = G(\cdot), \forall i, j.
\label{eq:arbit_objective}
\vspace{-5pt}
\end{equation}

After deriving $\hat{z}$, the flow model $f_\theta$ transforms $\hat{z}$ into an SR image $\hat{y}$, given by: $\hat{y} = f_\theta^{-1}(\hat{z}; \cdot)$.
Note that only the latent module undergoes training, the pre-trained flow model $f_\theta$ remains frozen during both training and inference phases.

\subsection{Framework Overview}
\vspace{-3pt}
\begin{figure*}[t]
  \centering
  \includegraphics[width=1\linewidth]{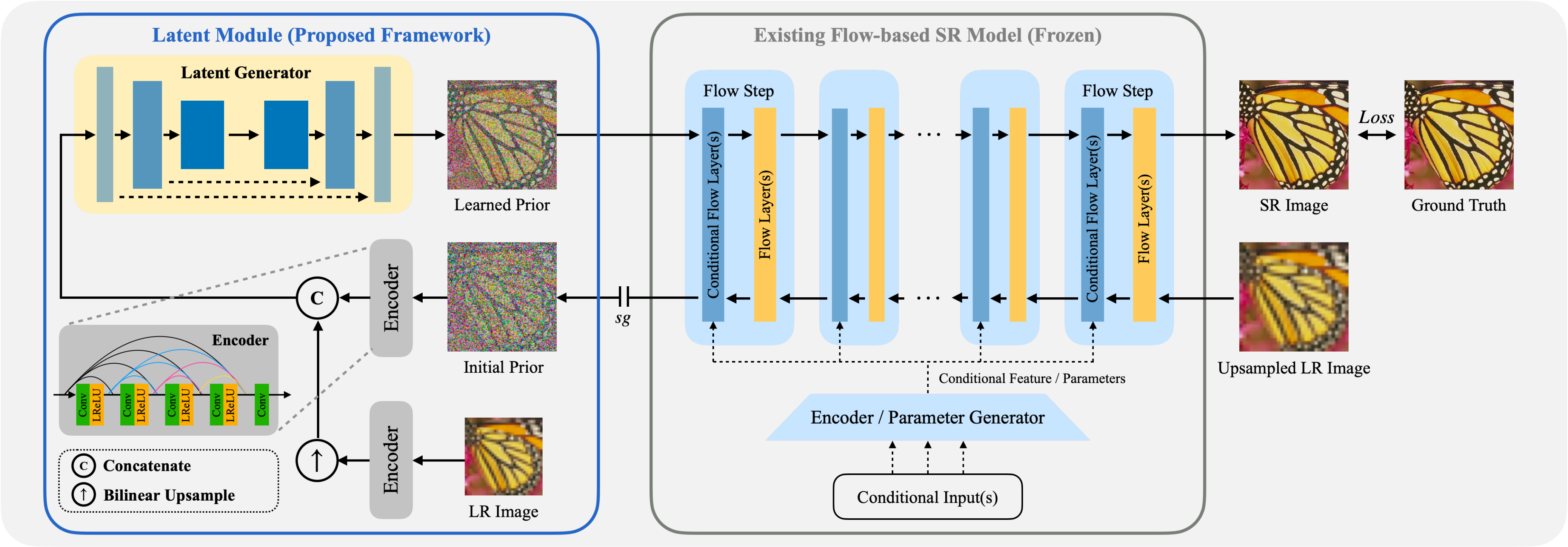}
  \caption{
    An overview of the proposed inference scheme for flow-based SR models. This framework consists of the proposed latent module and a flow model, which can be any existing flow-based SR model. The latent module generates a learned latent by leveraging the information extracted from an LR image. Then, the flow model transforms this latent into the corresponding SR image.
  }
  \label{fig:overview}
\vspace{-12pt}
\end{figure*}
\label{sec::overview}
Fig.~\ref{fig:overview} illustrates an overview of our proposed framework, which aims to leverage a conditional learned prior to address the inherent issues in flow-based SR models. Specifically, it consists of a proposed latent module and a flow model, with the latter being any contemporary flow-based SR model. The latent module processes signals from the LR image to produce a learned prior, which is then transformed by the flow model into the SR images. In this work, we integrate our framework with two existing flow models, including LINF~\cite{linf}, which represents arbitrary-scale SR framework, and SRFlow~\cite{srflow} for fixed-scale SR method.
\subsection{Latent Module}
\vspace{-3pt}
\label{sec::latent_module}
\paragraph{Overview.}
The latent module is designed to predict the conditional learned prior by utilizing the LR conditional signals, where this prior is a latent code that holds several properties capable of addressing the inherent issues of flow models, as mentioned in Section~\ref{sec::introduction}. To achieve this, the latent module is designed to: (1) 
fuse information across patches 
to mitigate discontinuities in image content (\ie, grid artifacts), (2) leverage LR conditional signals to avoid out-of-distribution predictions that lead to an exploding inverse, and (3) directly output a learned prior without random sampling.
As a result, this work introduces a latent module that extracts features of an LR image from both image space and latent space, and processes these features with a deep neural network to derive the conditional learned prior. Specifically, the architecture of the latent module comprises two feature encoders and a latent generator. One encoder processes the LR image to extract image space features, while the other works on the ``initial prior’’ to capture latent space signals, where the initial prior is the latent code corresponding to the upsampled LR image. Then, the latent generator utilizes these features to produce the learned latent code. The introduction of an initial prior provides a promising initialization in the latent space for seeking $z^*$, thus easing the prior generation process. A detailed analysis of this effect is presented in Section~\ref{sec::ablation_input}.
\vspace{-10pt}
\paragraph{The Design of the Latent Module.}
Regarding the design of the latent module, the feature encoders adopt a five-layer dense block~\cite{densenet} architecture but maintain independent weights to process inputs from different spaces. On the other hand, the architecture of the latent generator is highly flexible, which allows the incorporation of any commonly used module. In this work, UNet~\cite{unet} is primarily selected as the latent generator due to its efficiency and capabilities in capturing multi-level features, which aligns well with our objective. In addition, Section~\ref{sec::ablation_design} provides a detailed analysis of alternative architectures of the latent generator, including EDSR-baseline~\cite{edsr} and Swin Transformer~\cite{swin}.

\subsection{Objective Function}
\vspace{-3pt}
\label{sec::objective}
To generate a latent code $\hat{z}$ that approximates the optimal latent code $z^*$, our framework employs an objective function that aims at boosting both the accuracy in latent space and the perceptual quality of generated images. 
For the learning in latent space, we define a loss function $\mathcal{L}_{latent}$ to minimize the L1 distance between $\hat{z}$
and  $z^*$, formulated as:
\vspace{-7pt}
\begin{equation}
    \mathcal{L}_{latent} = \frac{1}{N} \sum_i^{N} ||z_i^* - \hat{z}_i||_1,
\label{eq:latent_loss}
\vspace{-7pt}
\end{equation}

where $z^*$ is obtained by transforming the HR image into the corresponding latent code with a flow model, which is available during training. In addition, we adopt the VGG perceptual loss~\cite{vggloss} $\mathcal{L}_{percep}$ in the image space to produce enhanced SR results. This objective guides our latent module to generate a latent code, which corresponds to an image that visually resembles the HR image. Specifically, it calculates the L1 distance between features extracted by the SR image and the HR image, with a pre-trained VGG19~\cite{vgg} network $\Psi_{per}$, which can be expressed as:
\vspace{-7pt}
\begin{equation}
    \mathcal{L}_{percep} = \frac{1}{N} \sum_i^{N} ||\Psi_{per}(y_i) - \Psi_{per}(\hat{y_i})||_1.
\label{eq:vgg_loss}
\vspace{-7pt}
\end{equation}

As a result, the overall objective can be expressed as:
\vspace{-8pt}
\begin{equation}
    \mathcal{L}_{total} = \mathcal{L}_{percep} + \lambda\mathcal{L}_{latent}.
\label{eq:total_loss}
\vspace{-8pt}
\end{equation}

In practice, $\lambda$ is set to 0 when integrating our framework with LINF, and 0.1 when combined with SRFlow based on the following observations. First, we find the sole application of Eq.~(\ref{eq:latent_loss}) results in an average prediction across all potential latent codes. This outcome resembles the ``regression-to-the-mean’’~\cite{mean} effect observed with L1 regression loss in image space.
Furthermore, while adopting the perceptual loss is sufficient when integrating our framework with LINF model, SRFlow experiences training instability under this scheme, where infinite values emerge occasionally. To address this phenomenon, we employ the latent space loss $\mathcal{L}_{latent}$  as a regularization term, which effectively prevents the generation of out-of-distribution latent codes~\cite{robust} that lead to an exploding inverse. Section~\ref{sec::ablation_loss} delivers an analysis of the impact of objective functions.
\vspace{-5pt}

\vspace{-2pt}
\section{Experiments}
\label{sec::experiments}
\vspace{-3pt}
In this section, we present the experimental results and ablation studies. The results demonstrate how the proposed framework addressing the inherent issues of flow-based SR methods, and the effects after integrating our proposed framework with LINF~\cite{linf} and SRFlow~\cite{srflow}.
\vspace{-3pt}
\subsection{Experimental Setup}
\label{sec::setup}
\vspace{-3pt}
\paragraph{Arbitrary-scale Flow-based SR.}
When integrating our framework with the arbitrary-scale framework LINF~\cite{linf}, we utilize two variants of their 3$\times$3 patch-based models: RRDB-LINF and EDSR-baseline-LINF. The former employs an RRDB~\cite{esrgan} as the image encoder, whereas the latter uses an EDSR-baseline~\cite{edsr} backbone. For RRDB-LINF, we adopt their released model. In the case of EDSR-baseline-LINF, we reproduced the baseline model using their codebase due to the absence of a pre-trained model. Upon incorporating our proposed ``\textit{Learned Prior}" (LP) into LINF, we refer to the enhanced versions as ``\textit{LINF-LP}", with two configurations: ``\textit{EDSR-baseline-LINF-LP}" and ``\textit{RRDB-LINF-LP}". To train LINF-LP, we set the LR image size to 96$\times$96,
and crop the corresponding 96$s$ $\times$ 96$s$ HR image patch as ground truth, where $s$ denotes a continuous upsampling scale $s \in \mathcal{U}(1, 4)$. LINF-LP is trained over 1,000 epochs with a batch size of 16. The initial learning rate is $1e^{-4}$, which is halved at [200, 400, 600, 800] epochs when using UNet and EDSR-baseline as the latent generator, and at [500, 850, 900, 950] epochs for Swin-T.
%
\vspace{-11pt}
\paragraph{Fixed-scale Flow-based SR.}
For the fixed-scale SR method SRFlow~\cite{srflow}, we also implement the enhanced version ``\textit{SRFlow-LP}". The experiments of SRFLow-LP are conducted on the DIV2K 4x SR task, and we integrate our proposed framework with their released pre-trained model. We train SRFlow-LP over five epochs, using a batch size of 12. The HR image size is set to 160$\times$160, and the initial learning rate is $1e^{-4}$, which is halved after each epoch.
\vspace{-11pt}
\paragraph{Datasets.}
We use the DIV2K~\cite{div2k} dataset for training EDSR-baseline-LINF-LP, and the joint of DIV2K and Flickr2K~\cite{flickr2k} to train RRDB-LINF-LP and SRFlow-LP. These two datasets consist of 800 and 2,650 images, respectively. The evaluation is conducted on the DIV2K validation set. For the arbitrary-scale framework LINF-LP, we further test on several SR benchmark datasets, including Set5~\cite{set5}, Set14~\cite{set14}, B100~\cite{b100}, and Urban100~\cite{urban100}.
\vspace{-11pt}
\paragraph{Network Details.}
For our proposed latent module, we utilize a three-layer deep UNet~\cite{unet} as the latent generator. It starts with an initial feature dimension of 64, comprising both initial prior and LR image features, each having a dimension of 32. The feature dimension is doubled with each downsample operation and halved with each upsample operation.
For the ablation study in Section~\ref{sec::ablation_design}, the feature dimensions of EDSR-baseline~\cite{edsr} and Swin Transformer (Swin-T)~\cite{swin} are 64 and 192, respectively.
\vspace{-11pt}
\paragraph{Evaluation Metrics.}
For evaluation, we select several commonly used metrics in SR tasks. These include PSNR and SSIM for fidelity measurements, and LPIPS~\cite{lpips} for assessing perceptual quality. In addition, we employ the LR-PSNR metric~\cite{ntire2021}, which calculates the PSNR between the bicubic downsampled SR image and the original LR image. 
%
\subsection{Experimental Results}
\vspace{-2pt}
\subsubsection{Generative Super-Resolution}
\vspace{-6pt}
We compare the performance of LINF-LP and SRFlow-LP with Flow-based~\cite{hcflow, srflow, linf} and other generative SR methods~\cite{esrgan, ranksrgan, srdiff, srooe}.
The results in Table~\ref{tab:generative_sr} reveal the benefits of our framework in 
two aspects: effectiveness and generalizability.
Firstly, both SRFlow-LP and LINF-LP demonstrate significant improvements in fidelity- and perception-oriented metrics. For instance, SRFlow-LP gains an improvement of 0.42 dB in PSNR and 0.012 in LPIPS, and LINF-LP achieves an improvement of 0.62 dB and 0.67 dB in PSNR, along with gains of 0.01 and 0.007 in LPIPS when using EDSR-baseline and RRDB backbones, respectively. 
%
Table~\ref{tab:generative_sr} also demonstrates that our proposed framework enables flow-based methods to achieve comparability with state-of-the-art methods~\cite{esrgan, ranksrgan, srdiff, srooe, hcflow}.
In addition, the simultaneous enhancements in both metrics present the capability of our framework to further push the boundary of the perception-distortion trade-off~\cite{tradeoff}.
Secondly, the improvements observed in both SRFlow-LP and LINF-LP exhibit the generalizability of our approach, which is capable of extending to both fixed-scale and arbitrary-scale frameworks without the need for customized components. This finding validates the capability of the proposed framework across various SR scenarios. 

\begin{table}[t]
\caption{The 4$\times$ SR results on the DIV2K~\cite{div2k} validation set. The best results are highlighted in \red{red}.}
\vspace{-7pt}
\renewcommand{\arraystretch}{1.3}
\newcommand{\mytoprule}{\toprule[1.2pt]}
\newcommand{\mybottomrule}{\bottomrule[1.2pt]}
\centering
\resizebox{1.0\columnwidth}{!}{%
        \begin{tabular}{l|c|c|c|c}
        \mybottomrule
        Generative SR Method & PSNR$(\uparrow)$ & SSIM$(\uparrow)$ & LPIPS$(\downarrow)$ & LR-PSNR$(\uparrow)$ \\
        \hline
        ESRGAN~\cite{esrgan} & 26.22 & 0.75 & 0.124 & 39.03 \\
        RankSRGAN~\cite{ranksrgan} & 26.55 & 0.75 & 0.128 & 42.33 \\
        SRDiff~\cite{srdiff} & 27.41 & \textbf{\textcolor{red}{0.79}} & 0.136 & \textbf{\textcolor{red}{55.21}} \\
        SROOE~\cite{srooe} & \textbf{\textcolor{red}{27.69}} & \textbf{\textcolor{red}{0.79}} & \textbf{\textcolor{red}{0.096}} & 50.80 \\
        \mytoprule
        \mybottomrule
        Flow-based Generative SR Method & PSNR$(\uparrow)$ & SSIM$(\uparrow)$ & LPIPS$(\downarrow)$ & LR-PSNR$(\uparrow)$ \\
        \hline
        HCFlow++ ($\tau = 0.9$)~\cite{hcflow} & 26.61 & 0.74 & 0.110 & 50.07 \\
        SRFlow ($\tau = 0.9$)~\cite{srflow} & 27.09 & 0.76 & 0.121 & 49.96 \\
        \textbf{SRFlow-LP (Ours)} & \textbf{\textcolor{red}{27.51}} & \textbf{\textcolor{red}{0.78}} & \textbf{\textcolor{red}{0.109}} & \textbf{\textcolor{red}{51.51}} \\
        \hline \hline
        EDSR-baseline-LINF ($\tau = 0.8$)~\cite{linf} & 27.02 & 0.76 & 0.130 & 43.19 \\
        \textbf{EDSR-baseline-LINF-LP (Ours)} & \textbf{\textcolor{red}{27.64}} & \textbf{\textcolor{red}{0.78}} & \textbf{\textcolor{red}{0.119}} & \textbf{\textcolor{red}{46.96}} \\
        \hline
        RRDB-LINF ($\tau = 0.8$)~\cite{linf} & 27.33 & 0.77 & 0.112 & 43.64 \\
        \textbf{RRDB-LINF-LP (Ours)} & \textbf{\textcolor{red}{28.00}} & \textbf{\textcolor{red}{0.78}} & \textbf{\textcolor{red}{0.105}} & \textbf{\textcolor{red}{47.30}} \\
        \mytoprule
        \end{tabular}
    }
\label{tab:generative_sr}
\vspace{-15pt}
\end{table}
%
\vspace{-10pt}
\subsubsection{Arbitrary-scale Flow-based SR}
\vspace{-5pt}
\paragraph{Quantitative Results.}
{
\begin{table*}[t]
\caption{The arbitrary-scale SR results on SR benchmark datasets. ``\textit{In-scales}'' and ``\textit{OOD-scales}'' refer to in- and out-of-training-distribution scales. LPIPS~\cite{lpips} scores are reported (lower is better), with the best and second-best highlighted in \textbf{\red{red}} and \blue{blue}, respectively.}
\vspace{-5pt}
\newcommand{\mytoprule}{\toprule[2.8pt]}
\newcommand{\mybottomrule}{\bottomrule[2.8pt]}
\renewcommand{\arraystretch}{1.3}
\centering
\setlength{\tabcolsep}{1.0em}
\resizebox{2.1\columnwidth}{!}{%
    \huge
    \begin{tabular}{l|ccc|cc|ccc|cc|ccc|cc|ccc|cc}
    \mybottomrule
     & \multicolumn{5}{c|}{Set5~\cite{set5}} & \multicolumn{5}{c|}{Set14~\cite{set14}} & \multicolumn{5}{c|}{B100~\cite{b100}} & \multicolumn{5}{c}{Urban100~\cite{urban100}} \\ \cline{2-21}
     
     Method & 
     \multicolumn{3}{c|}{In-scales} & \multicolumn{2}{c|}{OOD-scales} & \multicolumn{3}{c|}{In-scales} & \multicolumn{2}{c|}{OOD-scales} &
     \multicolumn{3}{c|}{In-scales} & \multicolumn{2}{c|}{OOD-scales} & \multicolumn{3}{c|}{In-scales} & \multicolumn{2}{c}{OOD-scales} \\
     
     & $\times$2 & $\times$3 & $\times$4 & $\times$6 & $\times$8
     & $\times$2 & $\times$3 & $\times$4 & $\times$6 & $\times$8
     & $\times$2 & $\times$3 & $\times$4 & $\times$6 & $\times$8
     & $\times$2 & $\times$3 & $\times$4 & $\times$6 & $\times$8 \\
     
     \hline
     
     EDSR-baseline-MetaSR~\cite{metasr} & 0.057 & 0.125 & 0.175 & 0.253 & 0.326 & 0.094 & 0.207 & 0.286 & 0.395 & 0.460 & 0.147 & 0.285 & 0.376 & 0.492 & 0.565 & 0.065 & 0.157 & 0.233 & 0.352 & 0.446 \\
     
     EDSR-baseline-LIIF~\cite{liif} & 0.056 & 0.124 & 0.173 & 0.248 & 0.307 & 0.093 & 0.205 & 0.284 & 0.390 & 0.449 & 0.147 & 0.282 & 0.372 & 0.486 & 0.556 & 0.064 & 0.155 & 0.228 & 0.338 & 0.422 \\
     
     EDSR-baseline-LTE~\cite{lte} & 0.056 & 0.123 & 0.174 & 0.257 & 0.326 & 0.092 & 0.203 & 0.283 & 0.396 & 0.463 & 0.146 & 0.280 & 0.371 & 0.495 & 0.570 & 0.063 & 0.152 & 0.224 & 0.345 & 0.436 \\
     
     EDSR-baseline-LINF~\cite{linf} ($\tau$ = $\tau_0$) & \textcolor{blue}{0.035} & \textcolor{blue}{0.067} & \textcolor{blue}{0.088} & \textcolor{blue}{0.158} & \textcolor{blue}{0.249} & \textcolor{blue}{0.064} & \textcolor{blue}{0.115} & \textcolor{blue}{0.163} & \textcolor{blue}{0.275} & \textcolor{blue}{0.375} & \textcolor{blue}{0.108} & \textcolor{blue}{0.172} & \textcolor{blue}{0.207} & \textcolor{blue}{0.319} & \textcolor{blue}{0.451} & \textcolor{blue}{0.050} & \textcolor{blue}{0.110} & \textcolor{blue}{0.158} & \textcolor{blue}{0.273} & \textcolor{blue}{0.386} \\
     
     \textbf{EDSR-baseline-LINF-LP (Ours)} & \textbf{\textcolor{red}{0.026}} & \textbf{\textcolor{red}{0.047}} & \textbf{\textcolor{red}{0.074}} & \textbf{\textcolor{red}{0.145}} & \textbf{\textcolor{red}{0.243}} & \textbf{\textcolor{red}{0.054}} & \textbf{\textcolor{red}{0.094}} & \textbf{\textcolor{red}{0.144}} & \textbf{\textcolor{red}{0.253}} & \textbf{\textcolor{red}{0.364}} & \textbf{\textcolor{red}{0.084}} & \textbf{\textcolor{red}{0.127}} & \textbf{\textcolor{red}{0.177}} & \textbf{\textcolor{red}{0.289}} & \textbf{\textcolor{red}{0.425}} & \textbf{\textcolor{red}{0.044}} & \textbf{\textcolor{red}{0.098}} & \textbf{\textcolor{red}{0.146}} & \textbf{\textcolor{red}{0.253}} & \textbf{\textcolor{red}{0.377}} \\
     
     \hline \hline
     
     RRDB-LINF~\cite{linf} ($\tau$ = $\tau_0$) & \textcolor{blue}{0.034} & \textcolor{blue}{0.064} & \textcolor{blue}{0.084} & \textcolor{blue}{0.147} & \textcolor{blue}{0.247} & \textcolor{blue}{0.059} & \textcolor{blue}{0.110} & \textcolor{blue}{0.146} & \textcolor{blue}{0.252} & \textcolor{blue}{0.359} & \textcolor{blue}{0.097} & \textcolor{blue}{0.152} & \textcolor{blue}{0.194} & \textcolor{blue}{0.306} & \textcolor{blue}{0.444} & \textcolor{blue}{0.040} & \textcolor{blue}{0.093} & \textcolor{blue}{0.137} & \textcolor{blue}{0.239} & \textcolor{blue}{0.354} \\
     
     \textbf{RRDB-LINF-LP (Ours)} & \textbf{\textcolor{red}{0.023}} & \textbf{\textcolor{red}{0.042}} & \textbf{\textcolor{red}{0.066}} & \textbf{\textcolor{red}{0.131}} & \textbf{\textcolor{red}{0.234}} & \textbf{\textcolor{red}{0.043}} & \textbf{\textcolor{red}{0.087}} & \textbf{\textcolor{red}{0.124}} & \textbf{\textcolor{red}{0.221}} & \textbf{\textcolor{red}{0.322}} & \textbf{\textcolor{red}{0.061}} & \textbf{\textcolor{red}{0.113}} & \textbf{\textcolor{red}{0.163}} & \textbf{\textcolor{red}{0.264}} & \textbf{\textcolor{red}{0.378}} & \textbf{\textcolor{red}{0.033}} & \textbf{\textcolor{red}{0.081}} & \textbf{\textcolor{red}{0.126}} & \textbf{\textcolor{red}{0.219}} & \textbf{\textcolor{red}{0.331}} \\
     
    \mytoprule
    \end{tabular}
}
\label{tab:exp:arbit_sr}
\vspace{-10pt}
\end{table*}
}
For the arbitrary-scale SR framework LINF~\cite{linf}, we compare the performance of EDSR-baseline-LINF-LP and RRDB-LINF-LP with their corresponding baselines.
We evaluate the performance at both in-distribution (2$\times$, 3$\times$, 4$\times$) and out-of-training-distribution (6$\times$, 8$\times$) upsampling scales on widely used SR benchmark datasets~\cite{set5, set14, b100, urban100}. Since our focus is flow-based generative models, we adopt LPIPS~\cite{lpips} for assessing the perceptual quality of generated images. In addition, LINF adopts a specific sampling temperature $\tau$ for different scaling factors: $\tau = 0.5$ for 2$\times$, 3$\times$, 4$\times$ SR, $\tau = 0.4$ for 6$\times$ SR, and $\tau = 0.2$ for 8$\times$ SR. In contrast, our framework directly predicts a latent code from the LR image without the need to fine-tune this hyperparameter.

Table~\ref{tab:exp:arbit_sr} demonstrates the adaptability and the potential
of our framework. 
(1) Both EDSR-baseline-LINF-LP and RRDB-LINF-LP achieve considerable improvements across in-distribution and out-of-training-distribution scales. This demonstrates the ability of our framework to adaptively predict a conditional learned prior for arbitrary upsampling scales, even at OOD scales. (2) RRDB-LINF-LP typically shows greater enhancement than EDSR-baseline-LINF-LP. For instance, RRDB-LINF-LP gains an improvement of 0.022 in LPIPS on Urban100 8$\times$ SR, compared to a 0.009 enhancement by EDSR-baseline-LINF-LP. We attribute this to the stronger capability of the RRDB~\cite{esrgan} backbone, which provides superior learning signals for the latent module. In light of this, we suggest the enhancements achieved by our framework are proportional to the capacity of the flow models, with more powerful models potentially yielding more pronounced improvements. 

\paragraph{Mitigating Grid Artifacts.}
\label{sec::grid}
\vspace{-10pt}
\begin{figure}[t]
  \centering
  \includegraphics[width=1\linewidth]{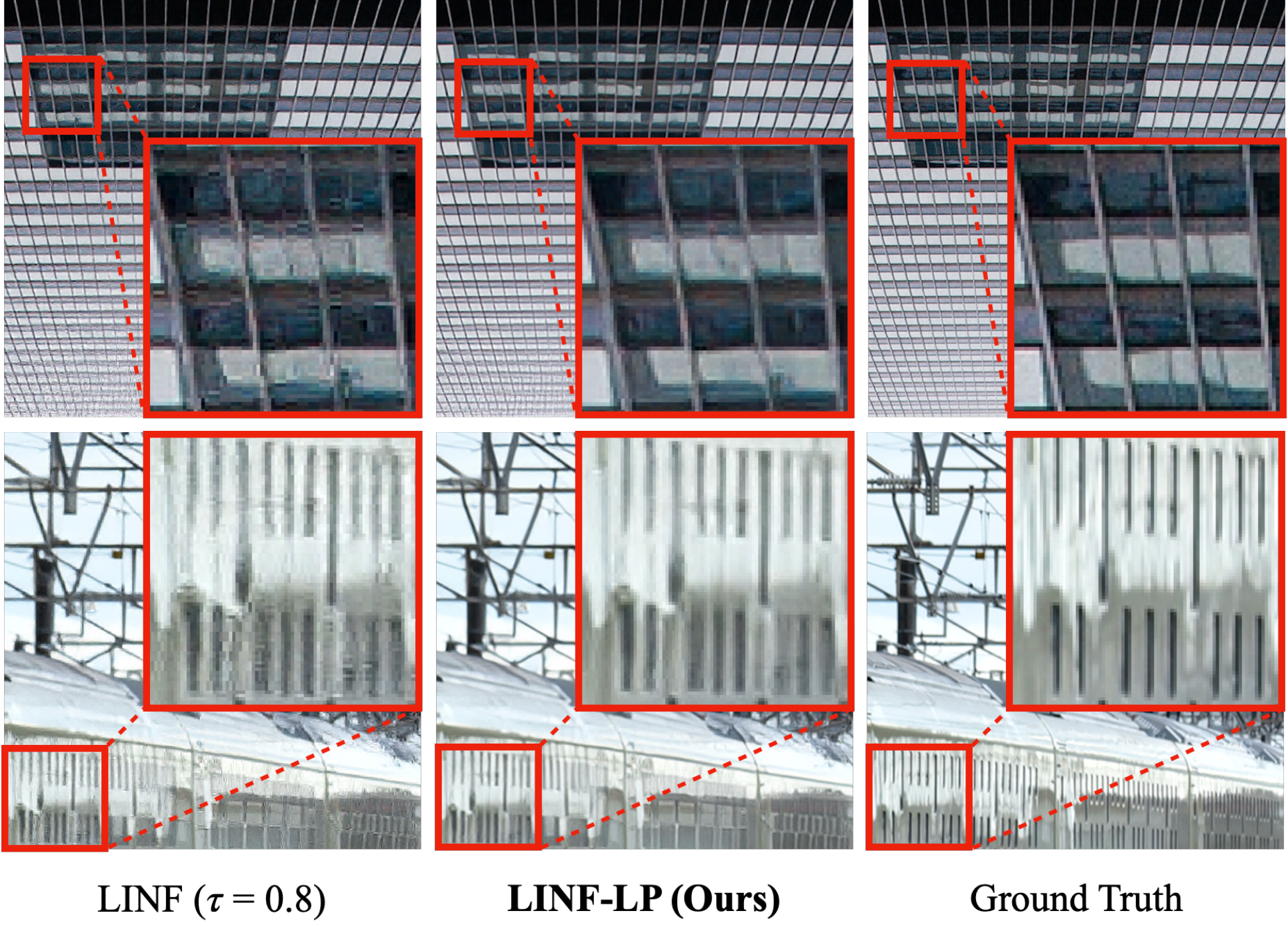}
  \vspace{-2em}
  \caption{
    A qualitative comparison between the 4$\times$ SR results of LINF~\cite{linf} and LINF-LP. (zoom in for better clarity)
  }
  \label{fig:grid}
  \vspace{-15pt}
\end{figure}
As illustrated in Fig.~\ref{fig:grid}, grid artifacts are prominent in images produced by LINF~\cite{linf}. This issue arises since LINF constructs an image by combining independently sampled patches. When utilizing a higher temperature $\tau$ to generate images with diverse content, the magnitude of values sampled by adjacent patches could vary greatly, which leads to discontinuities in image content. However, LINF-LP effectively reduces the presence of grid artifacts. The key to this improvement lies in the learned prior predicted by our latent module, which efficiently captures global information from an LR image and can guide LINF to generate coherent content.

%
\vspace{-10pt}
\subsubsection{Fixed-scale Flow-based SR}
\vspace{-5pt}
\paragraph{The Occurrence of Exploding Inverses.}
\label{sec::explode}
\begin{table}[t]
\caption{The 4$\times$ SR results on the DIV2K~\cite{div2k} validation set, which are the average of 10 validation runs. ``\textit{\%Inf}''~\cite{explode, robust} represents the probability of generating an exploding inverse. The best and second best results are highlighted in \textbf{\red{red}} and \blue{blue}, respectively.} 
\vspace{-5pt}
\renewcommand{\arraystretch}{1.4}
\newcommand{\mytoprule}{\toprule[1.3pt]}
\centering
\scriptsize
\resizebox{1\columnwidth}{!}{%
    \begin{tabular}{l|c|c|c}
    \bottomrule
    Method & \%Inf $(\downarrow)$ & LPIPS$(\downarrow)$ & 
    LR-PSNR$(\uparrow)$ \\
    \hline
    SRFlow ($\tau$ = 0.8)~\cite{srflow} & \textbf{\textcolor{red}{0}} & 0.124 & \textcolor{blue}{50.35} \\
    SRFlow ($\tau$ = 0.9)~\cite{srflow} & 0.8 & \textcolor{blue}{0.120} & 49.93 \\
    SRFlow ($\tau$ = 1.0)~\cite{srflow} & 6.7 & 0.130 & 48.62 \\
    \hline
    \textbf{SRFlow-LP (Ours)} & \textbf{\textcolor{red}{0}} & \textbf{\textcolor{red}{0.109}} & \textbf{\textcolor{red}{51.51}} \\
    \toprule
    \end{tabular}
}
\label{tab:inf}
\vspace{-12pt}
\end{table}
This analysis examines the likelihood of SRFlow~\cite{srflow} and SRFlow-LP encountering exploding inverses and measures the quality of generated images with LPIPS and LR-PSNR. 
Table~\ref{tab:inf} demonstrates that SRFlow-LP effectively prevents the occurrence of exploding inverses. This enhancement could be attributed to the conditional learned prior predicted by our framework, 
which avoids out-of-distribution predictions that lead to subsequent exploding inverses~\cite{robust}. A detailed analysis of this effect is presented in Section~\ref{sec::ablation_loss}.

\begin{figure}[t]
  \centering
  \includegraphics[width=1\linewidth]{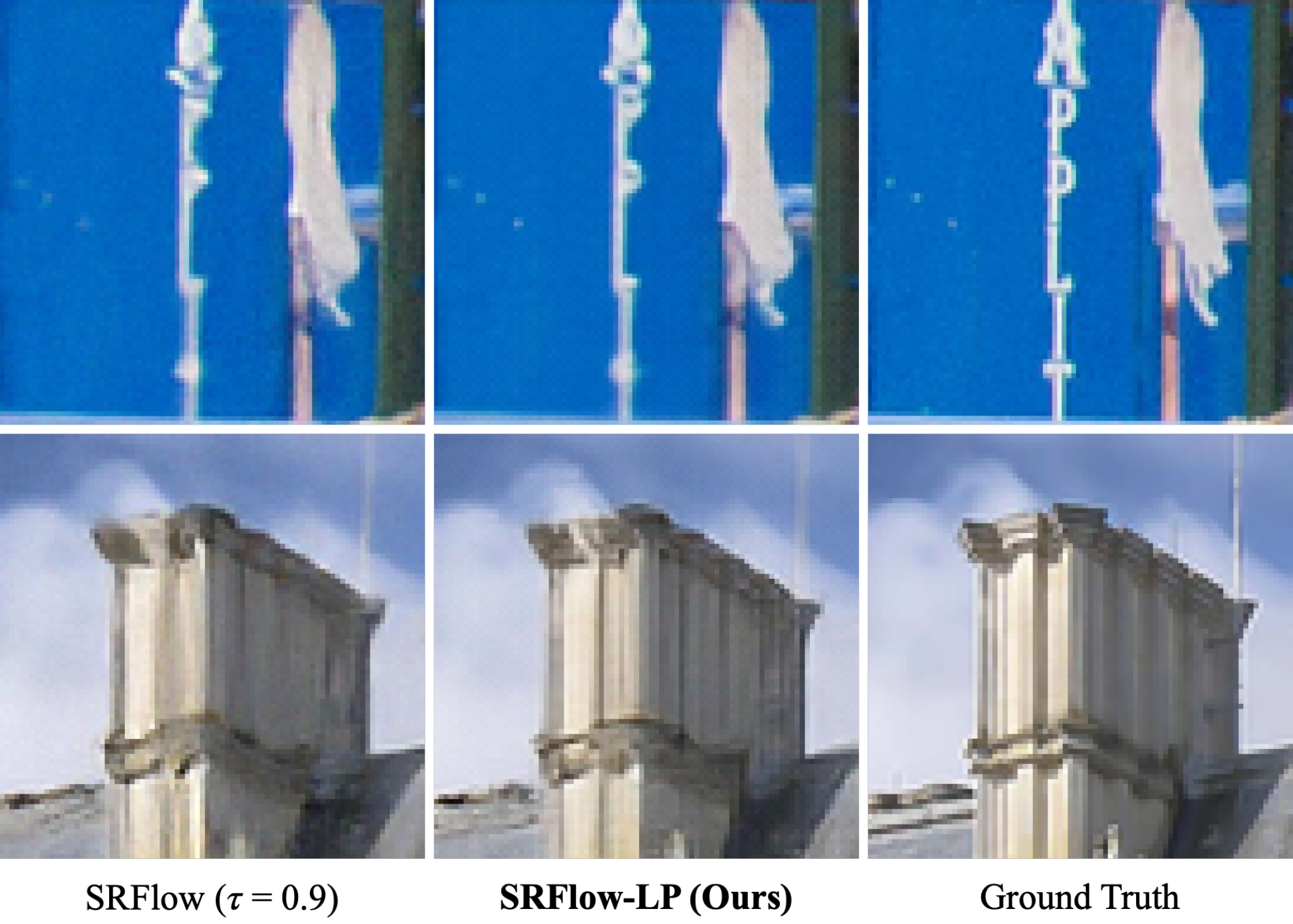}
  \vspace{-2em}
  \caption{
    A qualitative comparison between the 4$\times$ SR results of SRFlow~\cite{srflow} and SRFlow-LP.
  }
  \label{fig:srflow}
  \vspace{-10pt}
\end{figure}

For SRFlow, the frequency of producing exploding inverses rises with increasing temperature. To elaborate, SRFlow achieves optimal LPIPS scores at $\tau$ = 0.9, with a slight risk of encountering exploding inverses, which strikes a balance between perceptual quality and consistency. In addition, SRFlow shows distinct effects at $\tau$ = 0.8 and $\tau$ = 1.0. At a temperature $\tau$ = 0.8, SRFlow delivers a higher LR-PSNR score and shows no exploding inverses, while the perceptual quality is compromised. At $\tau$ = 1.0, the probability of encountering exploding inverses rises sharply. Also, excessively high temperatures could introduce image artifacts~\cite{linf}, thus affecting both LPIPS and LR-PSNR scores.
\paragraph{Qualitative Results.}
\vspace{-11pt}
Fig.~\ref{fig:srflow} presents a qualitative comparison between SRFlow-LP and SRFlow. The former can generate finer details such as lines and circles, while the latter struggles to render these even at a high-temperature setting of $\tau = 0.9$ for more diverse contents. This observation indicates the capabilities of our framework, which not only mitigates grid artifacts by integrating global information, as described in Section~\ref{sec::grid} but also excels in capturing intricate details, resulting in visually appealing effects.
\subsection{Ablation Studies}
\vspace{-3pt}
\label{sec::ablation}
This section presents ablation studies and in-depth discussions on the design of our latent generator, along with the effects of input selection and the objective function. In the second and third analyses, we adopt EDSR-baseline-LINF and EDSR-baseline-LINF-LP for comparison. For clarity, we denote them as ``LINF'' and ``LINF-LP'', respectively.
%
\paragraph{Design Flexibility of the Latent Generator.}
\vspace{-10pt}
\label{sec::ablation_design}
As described in Section~\ref{sec::latent_module}, the architecture of our latent generator allows the use of commonly used backbones. In this analysis, we explore this flexibility by adopting EDSR-baseline~\cite{edsr} and Swin-T~\cite{swin} as alternative latent generators. 
The results presented in Table~\ref{tab:ablation_design} illustrate the effectiveness and efficiency of our proposed framework. Firstly, LINF-LP yields considerable improvements with all these backbones in both fidelity- and perceptual-oriented metrics compared to LINF. Moreover, even the lightweight backbone EDSR-baseline, with a model size of only 1.4M, significantly boosts the performance. Among these results, LINF-LP (UNet) delivers the best PSNR and LPIPS scores. This superior performance could be attributed to the multi-level architecture of UNet~\cite{unet}, which effectively incorporates both local and global features into the learned prior.
\begin{table}[t]
\caption{The 4$\times$ SR results on the DIV2K~\cite{div2k} validation set. The names in the parentheses (\eg, UNet) refer to the architecture of our latent generator. The best and second-best results are highlighted in \textbf{\red{red}} and \blue{blue}, respectively.}
\vspace{-6pt}
\renewcommand{\arraystretch}{1.35}
\newcommand{\mytoprule}{\toprule[2.5pt]}
\newcommand{\mybottomrule}{\bottomrule[2.5pt]}
\centering
\footnotesize
\resizebox{1.0\columnwidth}{!}{%
    \huge
    \begin{tabular}{l|c|c|c|c}
    \mybottomrule
    Method & PSNR $(\uparrow)$ & SSIM $(\uparrow)$ & LPIPS $(\downarrow)$ & Parameters (M)  \\ 
    \hline
    EDSR-baseline-LINF (t=0.8)~\cite{linf} & 27.02 & 0.76 & 0.130 & 2.1 \\
    EDSR-baseline-LINF-LP (EDSR-baseline) & 27.53 & 0.77 & 0.121 & 2.1 + 1.4 \\
    EDSR-baseline-LINF-LP (UNet) & \textbf{\textcolor{red}{27.64}} & \textbf{\textcolor{red}{0.78}} & \textbf{\textcolor{red}{0.119}} & 2.1 + 4.6 \\
    EDSR-baseline-LINF-LP (Swin-T) & \textcolor{blue}{27.58} & \textbf{\textcolor{red}{0.78}} & \textcolor{blue}{0.120} & 2.1 + 7.3 \\
    \midrule
    \midrule
    RRDB-LINF (t=0.8)~\cite{linf} & 27.33 & 0.77 & 0.112 & 17.5 \\
    RRDB-LINF-LP (EDSR-baseline) & 27.86 & \textcolor{blue}{0.78} & 0.110 & 17.5 + 1.4 \\
    RRDB-LINF-LP (UNet) & \textbf{\textcolor{red}{28.00}} & \textcolor{blue}{0.78} & \textbf{\textcolor{red}{0.105}} & 17.5 + 4.6 \\
    RRDB-LINF-LP (Swin-T) & \textcolor{blue}{27.99} & \textbf{\textcolor{red}{0.79}} & \textbf{\textcolor{red}{0.105}} & 17.5 + 7.3 \\
    \mytoprule
    \end{tabular}
}
\label{tab:ablation_design}
\vspace{-6pt}
\end{table}

%
\paragraph{The Influence of Different Inputs.}
\vspace{-18pt}
\label{sec::ablation_input}
\begin{table}[t]
\caption{An analysis of the usage of different types of input. ``$\text{LPIPS}_{in}$'' denotes LPIPS~\cite{lpips} scores on the DIV2K 4$\times$ SR task, while ``$\text{LPIPS}_{OOD}$'' refers to LPIPS scores on the Urban100 8$\times$SR task. The best and second-best results are highlighted in \textbf{\red{red}} and \blue{blue}, respectively.}
\vspace{-6pt}
\renewcommand{\arraystretch}{1.25}
\newcommand{\mytoprule}{\toprule[1.2pt]}
\centering
\footnotesize
\resizebox{1\columnwidth}{!}{%
    \begin{tabular}{l|cc|c|c}
    \bottomrule
    Method & LR Image & Initial Prior & $\text{LPIPS}_{in}$ $(\downarrow)$ & $\text{LPIPS}_{OOD}$ $(\downarrow)$ \\
    \hline
    LINF~\cite{linf} & \xmark & \xmark & 0.130 & 0.386 \\
    LINF-LP (Ours) & \cmark & \xmark & 0.201 & 0.413 \\
    LINF-LP (Ours) & \xmark & \cmark & \textcolor{blue}{0.125} & \textcolor{blue}{0.384} \\
    \hline
    \textbf{LINF-LP (Ours)} & \cmark & \cmark & \textbf{\textcolor{red}{0.119}} & \textbf{\textcolor{red}{0.377}} \\
    
    \toprule
    \end{tabular}
}
\label{tab:ablation_input}
\vspace{-12pt}
\end{table}

We conduct analyses on the input to the proposed latent module by experimenting with various combinations, including (1) using the LR image only, (2) adopting the initial prior only, and (3) combining both, which is our final setting. During the analyses, we keep the total feature dimension at 64 in each experiment to ensure a fair comparison. 
The results in Table~\ref{tab:ablation_input} reveal the importance of both image space and latent space information. 
Firstly, an initial prior is crucial to our proposed framework. By solely adopting an initial prior as input, LINF-LP boosts the performance at both in- and out-of-distribution scales. This suggests an initial prior provides a promising initialization in latent space and eases the prior generation process.
Moreover, combining features from both image and latent spaces facilitates the best results, which infers that it is necessary to leverage both image and latent space features to enhance the quality of the learned prior. Lastly, relying solely on image space features is insufficient under our framework, as it is challenging to transform an LR image into a precise latent space signal with a single backbone. This setting leads to a decrease of 0.071 in LPIPS from the baseline on DIV2K 4$\times$ SR task.
%
\paragraph{Objective Function.}
\vspace{-10pt}
\label{sec::ablation_loss}
This analysis assesses the impact of different objective functions. The experiment employs three configurations: (1) using $\mathcal{L}_{latent}$ only, (2) adopting a combination of $\mathcal{L}_{latent}$ and $\mathcal{L}_{percep}$, and (3) exclusively adopting $\mathcal{L}_{percep}$, which is our chosen approach for LINF-LP. Note that in this analysis, the weight of $\mathcal{L}_{latent}$ is set to 0.1 when combined with $\mathcal{L}_{percep}$.
The results in Table~\ref{tab:ablation_loss} demonstrate that solely using $\mathcal{L}_{percep}$ yields the best LPIPS results. In contrast, employing $\mathcal{L}_{latent}$ only loss leads to an average prediction~\cite{mean} of all possible latent codes, resulting in inferior perceptual quality.
In addition, the dual application of $\mathcal{L}_{latent}$ and $\mathcal{L}_{percep}$ deteriorates LPIPS score by 0.014 at out-of-distribution (OOD) scales, yet slightly improves the in-distribution performance. 
This phenomenon suggests the $\mathcal{L}_{latent}$ enables LINF-LP to better fit the training distribution, as it receives guidance directly from the latent space (\ie, HR ground truth latent codes) during training, making its predictions toward in-distribution outcomes. Given that the performance at OOD scales is crucial to arbitrary-scale SR tasks, we only use $\mathcal{L}_{percep}$ for LINF-LP.

Based on the observation that the model trained with $\mathcal{L}_{latent}$ tends to yield latent codes within the training distribution, we adopt $\mathcal{L}_{latent}$ as a regularization term for SRFlow-LP. This approach aims to prevent SRFlow-LP from generating OOD predictions, which could lead to an exploding inverse as noted in~\cite{robust}. Under this setting, as illustrated in Table~\ref{tab:inf}, we successfully prevent the occurrence of exploding inverses without modifying the architecture~\cite{robust} and the pre-trained weights of SRFlow.
\begin{table}[t]
\caption{An analysis of the impact of the objective function. LPIPS and $\text{LPIPS}_{OOD}$ represent the LPIPS scores on the DIV2K 4$\times$SR and Urban100 8$\times$SR tasks, respectively. The best results are highlighted in \textbf{\red{red}}, with the second-best in \blue{blue}.}
\vspace{-5pt}
\renewcommand{\arraystretch}{1.25}
\newcommand{\mytoprule}{\toprule[1.3pt]}
\centering
\footnotesize
\resizebox{1\columnwidth}{!}{%
    \begin{tabular}{l|cc|c|c}
    \bottomrule
    Method & $\mathcal{L}_{latent}$ & $\mathcal{L}_{percep}$ & LPIPS $(\downarrow)$ & $\text{LPIPS}_{OOD}$ $(\downarrow)$ \\
    \hline
    LINF~\cite{linf} & \xmark & \xmark & 0.1297 & \textcolor{blue}{0.3863} \\
    LINF-LP (Ours) & \cmark & \xmark & 0.2658 & 0.4180 \\
    LINF-LP (Ours) & \cmark & \cmark & \textbf{\textcolor{red}{0.1185}} & 0.3914 \\
    \hline
    \textbf{LINF-LP (Ours)} & \xmark & \cmark & \textcolor{blue}{0.1191} & \textbf{\textcolor{red}{0.3774}} \\
    
    \toprule
    \end{tabular}
}
\label{tab:ablation_loss}
\vspace{-15pt}
\end{table}
\vspace{-5pt}

\section{Conclusion}
\label{sec::conclusions}
\vspace{-5pt}
In this work, we identify several challenges in flow-based SR methods, including grid artifacts, exploding inverses, and suboptimal results due to a fixed sampling temperature. To tackle these issues, we introduce a learned prior, which is predicted by the proposed latent module, to the inference phase of flow-based SR models.
This framework not only addresses the inherent issues in flow-based SR models but also enhances the performance of these models without modifying their original design or pre-trained weights. Our proposed framework is effective, flexible in design, and able to generalize to both fixed-scale and arbitrary-scale SR frameworks without requiring customized components.
\vspace{-3pt}

\vspace{-3pt}
\section*{Acknowledgments}
\vspace{-5pt}
The authors gratefully acknowledge the support from the National Science and Technology Council (NSTC) in Taiwan under grant numbers MOST 111-2223-E-007-004-MY3, as well as the financial support from MediaTek Inc., Taiwan. The authors would also like to express their appreciation for the donation of the GPUs from NVIDIA Corporation and NVIDIA AI Technology Center (NVAITC) used in this work. Furthermore, the authors extend their gratitude to the National Center for High-Performance Computing (NCHC) for providing the necessary computational and storage resources.

{
    \small
    \bibliographystyle{ieeenat_fullname}
    \bibliography{main}

\newcommand{\IJCV}{Int. J. Computer Vision (IJCV)}\newcommand{\CVPR}{Proc. IEEE Conf. on Computer Vision and Pattern Recognition (CVPR)}\newcommand{\CVPRW}{Proc. IEEE Conf. on Computer Vision and Pattern Recognition Workshop (CVPRW)}\newcommand{\ICCV}{Proc. IEEE Int. Conf. on Computer Vision (ICCV)}\newcommand{\ICCVW}{Proc. IEEE Int. Conf. on Computer Vision Workshop (ICCVW)}\newcommand{\ECCV}{Proc. European Conf. on Computer Vision (ECCV)}\newcommand{\ECCVW}{Proc. European Conf. on Computer Vision Workshop (ECCVW)}\newcommand{\IROS}{Proc. IEEE Int. Conf. on Intelligent Robots and Systems (IROS)}\newcommand{\CoRL}{Proc. Conf. on Robot Learning (CoRL)}\newcommand{\ICRA}{Proc. IEEE Int. Conf. on Robotics and Automation (ICRA)}\newcommand{\AAAI}{Proc. AAAI Conf. on Artificial Intelligence (AAAI)}\newcommand{\IJCAI}{Proc. Int. Joint Conf. on Artificial Intelligence (IJCAI)}\newcommand{\PAMI}{IEEE Trans. Pattern Analysis and Machine Intelligence (TPAMI)}\newcommand{\NIPS}{Proc. Conf. on Neural Information
  Processing Systems (NeurIPS)}\newcommand{\ICML}{Proc. Int. Conf. on Machine Learning (ICML)}\newcommand{\ICLR}{Proc. Int. Conf. on Learning Representations (ICLR)}\newcommand{\ICLRW}{Proc. Int. Conf. on Learning Representations Workshop (ICLRW)}\newcommand{\ICASSP}{Proc. IEEE Int. Conf. on Acoustics, Speech, & Signal Processing (ICASSP)}\newcommand{\BMVC}{Proc. British Machine Vision Conf. (BMVC)}\newcommand{\ACCV}{Proc. Asian Conf. on Computer Vision (ACCV)}\newcommand{\WACV}{Proc. IEEE Winter Conf. on Applications of Computer Vision (WACV)}
\begin{thebibliography}{82}
\providecommand{\natexlab}[1]{#1}
\providecommand{\url}[1]{\texttt{#1}}
\expandafter\ifx\csname urlstyle\endcsname\relax
  \providecommand{\doi}[1]{doi: #1}\else
  \providecommand{\doi}{doi: \begingroup \urlstyle{rm}\Url}\fi

\bibitem[Agustsson and Timofte(2017)]{div2k}
E. Agustsson and R. Timofte.
\newblock {NTIRE} 2017 challenge on single image super-resolution: Dataset and study.
\newblock In \emph{\CVPRW{}}, pages 1122--1131, 2017.

\bibitem[Ahn et~al.(2018)Ahn, Kang, and Sohn]{carn}
N. Ahn, B. Kang, and K.-A. Sohn.
\newblock Fast, accurate, and lightweight super-resolution with cascading residual network.
\newblock In \emph{\ECCV}, pages 252--268, 2018.

\bibitem[Andreas et~al.(2021)Andreas, Martin, and Radu]{ntire2021}
L. Andreas, D. Martin, and T. Radu.
\newblock Ntire 2021 learning the super-resolution space challenge.
\newblock In \emph{\CVPR}, pages 596--612, 2021.

\bibitem[Behrmann et~al.(2021)Behrmann, Vicol, Wang, Grosse, and Jacobsen]{explode}
J. Behrmann, P. Vicol, K.-C. Wang, R. Grosse, and J.-H. Jacobsen.
\newblock Understanding and mitigating exploding inverses in invertible neural networks.
\newblock In \emph{International Conference on Artificial Intelligence and Statistics}, pages 1792--1800. PMLR, 2021.

\bibitem[Bevilacqua et~al.(2012)Bevilacqua, Roumy, Guillemot, and Morel]{set5}
M. Bevilacqua, A. Roumy, C. Guillemot, and M.-L.~Alberi Morel.
\newblock Low-complexity single-image super-resolution based on nonnegative neighbor embedding.
\newblock In \emph{\BMVC{}}, pages 1--10, 2012.

\bibitem[Blau and Michaeli(2018)]{tradeoff}
Y. Blau and T. Michaeli.
\newblock The perception-distortion tradeoff.
\newblock In \emph{\CVPR}, 2018.

\bibitem[Bond-Taylor et~al.(2021)Bond-Taylor, Leach, Long, and Willcocks]{bond2021deep}
S. Bond-Taylor, A. Leach, Y. Long, and C.~G. Willcocks.
\newblock Deep generative modelling: A comparative review of vaes, gans, normalizing flows, energy-based and autoregressive models.
\newblock \emph{\PAMI}, 2021.

\bibitem[Bruna et~al.(2015)Bruna, Sprechmann, and LeCun]{mean}
J. Bruna, P. Sprechmann, and Y. LeCun.
\newblock Super-resolution with deep convolutional sufficient statistics.
\newblock \emph{arXiv preprint arXiv:1511.05666}, 2015.

\bibitem[C et~al.(2016)C, Loy, and Tang]{fsrcnn}
Dong C, C.~C. Loy, and X. Tang.
\newblock Accelerating the super-resolution convolutional neural network.
\newblock In \emph{\ECCV}, pages 391--407, 2016.

\bibitem[Cao et~al.(2023)Cao, Wang, Xian, Li, Ni, Pi, Zhang, Zhang, Timofte, and Gool]{ciaosr}
J. Cao, Q. Wang, Y. Xian, Y. Li, B. Ni, Z. Pi, K. Zhang, Y. Zhang, R. Timofte, and L.~Van Gool.
\newblock Ciaosr: Continuous implicit attention-in-attention network for arbitrary-scale image super-resolution.
\newblock In \emph{\CVPR}, pages 1796--1807, 2023.

\bibitem[Chadebec et~al.(2022)Chadebec, Vincent, and Allassonni{\`e}re]{chadebec2022pythae}
C. Chadebec, L. Vincent, and S. Allassonni{\`e}re.
\newblock Pythae: Unifying generative autoencoders in python-a benchmarking use case.
\newblock \emph{\NIPS}, 35:\penalty0 21575--21589, 2022.

\bibitem[Chen et~al.(2023{\natexlab{a}})Chen, Xu, Hong, Tsai, Kuo, and Lee]{clit}
H.-W. Chen, Y.-S. Xu, M.-F. Hong, Y.-M. Tsai, H.-K. Kuo, and C.-Y. Lee.
\newblock Cascaded local implicit transformer for arbitrary-scale super-resolution.
\newblock In \emph{\CVPR}, pages 18257--18267, 2023{\natexlab{a}}.

\bibitem[Chen et~al.(2017)Chen, Kingma, Salimans, Duan, Dhariwal, Schulman, Sutskever, and Abbeel]{chen2016variational}
X. Chen, D.~P. Kingma, T. Salimans, Y. Duan, P. Dhariwal, J. Schulman, I. Sutskever, and P. Abbeel.
\newblock Variational lossy autoencoder.
\newblock \emph{\ICLR}, 2017.

\bibitem[Chen et~al.(2023{\natexlab{b}})Chen, Wang, Zhou, Qiao, and Dong]{hat}
X. Chen, X. Wang, J. Zhou, Y. Qiao, and C. Dong.
\newblock Activating more pixels in image super-resolution transformer.
\newblock In \emph{\CVPR}, pages 22367--22377, 2023{\natexlab{b}}.

\bibitem[Chen et~al.(2021)Chen, Liu, and Wang]{liif}
Y. Chen, S. Liu, and X. Wang.
\newblock Learning continuous image representation with local implicit image function.
\newblock In \emph{\CVPR}, pages 8628--8638, 2021.

\bibitem[Dilokthanakul et~al.(2017)Dilokthanakul, Mediano, Garnelo, Lee, Salimbeni, Arulkumaran, and Shanahan]{dilokthanakul2016deep}
N. Dilokthanakul, P.~A.~M. Mediano, M. Garnelo, M.~C.~H. Lee, H. Salimbeni, K. Arulkumaran, and M. Shanahan.
\newblock Deep unsupervised clustering with gaussian mixture variational autoencoders.
\newblock \emph{\ICLR}, 2017.

\bibitem[Dong et~al.(2016)Dong, Loy, He, and Tang]{srcnn}
C. Dong, C.~C. Loy, K. He, and X. Tang.
\newblock Image super-resolution using deep convolutional networks.
\newblock \emph{\PAMI}, 38\penalty0 (2):\penalty0 295--307, 2016.

\bibitem[Dong et~al.(2022)Dong, Hangel, Chen, Sun, Bogner, Widhalm, You, Onofrey, de~Graaf, and Duncan]{mrsiflow}
S. Dong, G. Hangel, E.~Z. Chen, S. Sun, W. Bogner, G. Widhalm, C. You, J.~A. Onofrey, R. de Graaf, and J.~S. Duncan.
\newblock Flow-based visual quality enhancer for super-resolution magnetic resonance spectroscopic imaging.
\newblock In \emph{MICCAI Workshop on Deep Generative Models}, pages 3--13, 2022.

\bibitem[Esser et~al.(2021)Esser, Rombach, and Ommer]{esser2021taming}
P. Esser, R. Rombach, and B. Ommer.
\newblock Taming transformers for high-resolution image synthesis.
\newblock In \emph{\CVPR}, pages 12873--12883, 2021.

\bibitem[Ghosh et~al.(2020)Ghosh, Sajjadi, Vergari, Black, and Scholkopf]{ghosh2020from}
P. Ghosh, M.~S.~M. Sajjadi, A. Vergari, M. Black, and B. Scholkopf.
\newblock From variational to deterministic autoencoders.
\newblock In \emph{\ICLR}, 2020.

\bibitem[Goodfellow et~al.(2014)Goodfellow, Pouget-Abadie, Mirza, Xu, Warde-Farley, Ozair, Courville, and Bengio]{gan}
I. Goodfellow, J. Pouget-Abadie, M. Mirza, B. Xu, D. Warde-Farley, S. Ozair, A. Courville, and Y. Bengio.
\newblock Generative adversarial nets.
\newblock \emph{\NIPS}, 27, 2014.

\bibitem[Gulrajani et~al.(2017)Gulrajani, Kumar, Ahmed, Taiga, Visin, Vazquez, and Courville]{gulrajani2016pixelvae}
I. Gulrajani, K. Kumar, F. Ahmed, A.~A. Taiga, F. Visin, D. Vazquez, and A. Courville.
\newblock Pixelvae: A latent variable model for natural images.
\newblock \emph{\ICLR}, 2017.

\bibitem[Guo et~al.(2022)Guo, Zhang, Wu, Wang, Zhang, and Wang]{larsr}
B. Guo, X. Zhang, H. Wu, Y. Wang, Y. Zhang, and Y.-F. Wang.
\newblock Lar-sr: A local autoregressive model for image super-resolution.
\newblock In \emph{\CVPR}, pages 1899--1908, 2022.

\bibitem[Haris et~al.(2018)Haris, Shakhnarovich, and Ukita]{dbpn}
M. Haris, G. Shakhnarovich, and N. Ukita.
\newblock Deep back-projection networks for super-resolution.
\newblock In \emph{\CVPR}, pages 1664--1673, 2018.

\bibitem[Ho et~al.(2020)Ho, Jain, and Abbeel]{ho2020denoising}
J. Ho, A. Jain, and P. Abbeel.
\newblock Denoising diffusion probabilistic models.
\newblock \emph{\NIPS}, 33:\penalty0 6840--6851, 2020.

\bibitem[Hong et~al.(2023)Hong, Park, and Chun]{robust}
S. Hong, I. Park, and S.~Y. Chun.
\newblock On the robustness of normalizing flows for inverse problems in imaging.
\newblock In \emph{\ICCV}, pages 10745--10755, 2023.

\bibitem[Hu et~al.(2019)Hu, Mu, Zhang, Wang, Tan, and Sun]{metasr}
X. Hu, H. Mu, X. Zhang, Z. Wang, T. Tan, and J. Sun.
\newblock Meta-sr: {A} magnification-arbitrary network for super-resolution.
\newblock In \emph{\CVPR}, pages 1575--1584, 2019.

\bibitem[Huang et~al.(2017)Huang, Liu, Maaten, and Weinberger]{densenet}
G. Huang, Z. Liu, L.~Van~Der Maaten, and K.~Q. Weinberger.
\newblock Densely connected convolutional networks.
\newblock In \emph{\CVPR}, 2017.

\bibitem[Huang et~al.(2015)Huang, Singh, and Ahuja]{urban100}
J.-B. Huang, A. Singh, and N. Ahuja.
\newblock Single image super-resolution from transformed self-exemplars.
\newblock In \emph{\CVPR{}}, pages 5197--5206, 2015.

\bibitem[Johnson et~al.(2016)Johnson, Alahi, and Li]{vggloss}
J. Johnson, A. Alahi, and F.-F. Li.
\newblock Perceptual losses for real-time style transfer and super-resolution.
\newblock In \emph{\ECCV}, pages 694--711, 2016.

\bibitem[Karen and Andrew(2014)]{vgg}
S. Karen and Z. Andrew.
\newblock Very deep convolutional networks for large-scale image recognition.
\newblock \emph{arXiv preprint arXiv:1409.1556}, 2014.

\bibitem[Karras et~al.(2019)Karras, Laine, and Aila]{karras2019style}
T. Karras, S. Laine, and T. Aila.
\newblock A style-based generator architecture for generative adversarial networks.
\newblock In \emph{\CVPR}, pages 4401--4410, 2019.

\bibitem[Kim et~al.(2016)Kim, Lee, and Lee]{vdsr}
J. Kim, J.~K. Lee, and K.~M. Lee.
\newblock Accurate image super-resolution using very deep convolutional networks.
\newblock In \emph{\CVPR}, pages 1646--1654, 2016.

\bibitem[Kim and Son(2021)]{ncsr}
Y. Kim and D. Son.
\newblock Noise conditional flow model for learning the super-resolution space.
\newblock In \emph{\CVPR}, pages 424--432, 2021.

\bibitem[Kingma and Welling(2013)]{vae}
D.~P. Kingma and M. Welling.
\newblock Auto-encoding variational bayes.
\newblock \emph{arXiv preprint arXiv:1312.6114}, 2013.

\bibitem[Kingma et~al.(2016)Kingma, Salimans, Jozefowicz, Chen, Sutskever, and Welling]{kingma2016improved}
D.~P. Kingma, T. Salimans, R. Jozefowicz, X. Chen, I. Sutskever, and M. Welling.
\newblock Improved variational inference with inverse autoregressive flow.
\newblock \emph{\NIPS}, 29, 2016.

\bibitem[Ko et~al.(2023)Ko, Lee, Hong, Kim, and Ko]{mriflow}
K. Ko, B. Lee, J. Hong, D. Kim, and H. Ko.
\newblock Mriflow: Magnetic resonance image super-resolution based on normalizing flow and frequency prior.
\newblock \emph{Journal of Magnetic Resonance}, 352:\penalty0 107477, 2023.

\bibitem[Kobyzev et~al.(2020)Kobyzev, Prince, and Brubaker]{kobyzev2020normalizing}
I. Kobyzev, S.~J.~D. Prince, and M.~A. Brubaker.
\newblock Normalizing flows: An introduction and review of current methods.
\newblock \emph{\PAMI}, pages 3964--3979, 2020.

\bibitem[Lai et~al.(2017)Lai, Huang, Ahuja, and Yang]{lapsrn}
W.-S. Lai, J.-B. Huang, N. Ahuja, and M.-H. Yang.
\newblock Deep laplacian pyramid networks for fast and accurate super-resolution.
\newblock In \emph{\CVPR}, pages 624--632, 2017.

\bibitem[Ledig et~al.(2017)Ledig, Theis, Huszar, Caballero, Cunningham, Acosta, Aitken, Tejani, Totz, Wang, and Shi]{srgan}
C. Ledig, L. Theis, F. Huszar, J. Caballero, A. Cunningham, A. Acosta, A. Aitken, A. Tejani, J. Totz, Z. Wang, and W. Shi.
\newblock Photo-realistic single image super-resolution using a generative adversarial network.
\newblock In \emph{\CVPR}, pages 105--114, 2017.

\bibitem[Lee and Jin(2022)]{lte}
J. Lee and K.~H. Jin.
\newblock Local texture estimator for implicit representation function.
\newblock In \emph{\CVPR}, pages 1929--1938, 2022.

\bibitem[Li et~al.(2023)Li, Zhang, Liu, and Zhu]{craft}
A. Li, L. Zhang, Y. Liu, and C. Zhu.
\newblock Feature modulation transformer: Cross-refinement of global representation via high-frequency prior for image super-resolution.
\newblock In \emph{\ICCV}, pages 12514--12524, 2023.

\bibitem[Li et~al.(2022)Li, Yang, Chang, Feng, Xu, Li, and Chen]{srdiff}
H. Li, Y. Yang, M. Chang, H. Feng, Z. Xu, Q. Li, and Y. Chen.
\newblock Srdiff: Single image super-resolution with diffusion probabilistic models.
\newblock \emph{Neurocomputing}, 479:\penalty0 47--59, 2022.

\bibitem[Liang et~al.(2021{\natexlab{a}})Liang, Cao, Sun, Zhang, Gool, and Timofte]{swinir}
J. Liang, J. Cao, G. Sun, K. Zhang, L.~Van Gool, and R. Timofte.
\newblock Swinir: Image restoration using swin transformer.
\newblock In \emph{\ICCVW}, pages 1833--1844, 2021{\natexlab{a}}.

\bibitem[Liang et~al.(2021{\natexlab{b}})Liang, Lugmayr, Zhang, Danelljan, Gool, and Timofte]{hcflow}
J. Liang, A. Lugmayr, K. Zhang, M. Danelljan, L.~Van Gool, and R. Timofte.
\newblock Hierarchical conditional flow: A unified framework for image super-resolution and image rescaling.
\newblock In \emph{\ICCV{}}, pages 4056--4065, 2021{\natexlab{b}}.

\bibitem[Liang et~al.(2021{\natexlab{c}})Liang, Zhang, Gu, Gool, and Timofte]{fkp}
J. Liang, K. Zhang, S. Gu, L.~Van Gool, and R. Timofte.
\newblock Flow-based kernel prior with application to blind super-resolution.
\newblock In \emph{\CVPR}, pages 10601--10610, 2021{\natexlab{c}}.

\bibitem[Liang et~al.(2022)Liang, Zeng, and Zhang]{details}
Jie Liang, Hui Zeng, and Lei Zhang.
\newblock Details or artifacts: A locally discriminative learning approach to realistic image super-resolution.
\newblock In \emph{\CVPR}, pages 5657--5666, 2022.

\bibitem[Lim et~al.(2017)Lim, Son, Kim, Nah, and Lee]{edsr}
B. Lim, S. Son, H. Kim, S. Nah, and K.~M. Lee.
\newblock Enhanced deep residual networks for single image super-resolution.
\newblock In \emph{\CVPRW}, pages 1132--1140, 2017.

\bibitem[Liu et~al.(2021)Liu, Lin, Cao, Hu, Wei, Zhang, Lin, and Guo]{swin}
Z. Liu, Y. Lin, Y. Cao, H. Hu, Y. Wei, Z. Zhang, S. Lin, and B. Guo.
\newblock Swin transformer: Hierarchical vision transformer using shifted windows.
\newblock In \emph{\ICCV}, pages 10012--10022, 2021.

\bibitem[Lugmayr et~al.(2020)Lugmayr, Danelljan, Gool, and Timofte]{srflow}
A. Lugmayr, M. Danelljan, L.~Van Gool, and R. Timofte.
\newblock Srflow: Learning the super-resolution space with normalizing flow.
\newblock In \emph{\ECCV{}}, 2020.

\bibitem[Lugmayr et~al.(2022)Lugmayr, Danelljan, Yu, Gool, and Timofte]{adflow}
A. Lugmayr, M. Danelljan, F. Yu, L.~Van Gool, and R. Timofte.
\newblock Normalizing flow as a flexible fidelity objective for photo-realistic super-resolution.
\newblock In \emph{\WACV{}}, pages 874--883, 2022.

\bibitem[Martin et~al.(2001)Martin, Fowlkes, Tal, and Malik]{b100}
D. Martin, C. Fowlkes, D. Tal, and J. Malik.
\newblock A database of human segmented natural images and its application to evaluating segmentation algorithms and measuring ecological statistics.
\newblock In \emph{\ICCV{}}, pages 416--425, 2001.

\bibitem[Odena et~al.(2016)Odena, Dumoulin, and Olah]{deconv}
A. Odena, V. Dumoulin, and C. Olah.
\newblock Deconvolution and checkerboard artifacts.
\newblock \emph{Distill}, 2016.

\bibitem[Oord et~al.(2017)Oord, Vinyals, and Kavukcuoglu]{van2017neural}
A.~Van~Den Oord, O. Vinyals, and K. Kavukcuoglu.
\newblock Neural discrete representation learning.
\newblock \emph{\NIPS}, 30, 2017.

\bibitem[Razavi et~al.(2019)Razavi, den Oord, and Vinyals]{razavi2019generating}
A. Razavi, A.~Van den Oord, and O. Vinyals.
\newblock Generating diverse high-fidelity images with vq-vae-2.
\newblock \emph{\NIPS}, 32, 2019.

\bibitem[Rezende and Mohamed(2015)]{rezende2015variational}
D. Rezende and S. Mohamed.
\newblock Variational inference with normalizing flows.
\newblock In \emph{\ICML}, pages 1530--1538, 2015.

\bibitem[Ronneberger et~al.(2015)Ronneberger, Fischer, and Brox]{unet}
O. Ronneberger, P. Fischer, and T. Brox.
\newblock U-net: Convolutional networks for biomedical image segmentation.
\newblock In \emph{Medical Image Computing and Computer-Assisted Intervention}, pages 234--241, 2015.

\bibitem[S.-H. et~al.(2022)S.-H., Y.-S., and N.-I.]{fxsr}
Park S.-H., Moon Y.-S., and Cho N.-I.
\newblock Flexible style image super-resolution using conditional objective.
\newblock \emph{IEEE Access}, 10:\penalty0 9774--9792, 2022.

\bibitem[S.-H. et~al.(2023)S.-H., Y.-S., and N.-I.]{srooe}
Park S.-H., Moon Y.-S., and Cho N.-I.
\newblock Perception-oriented single image super-resolution using optimal objective estimation.
\newblock In \emph{\CVPR}, pages 1725--1735, 2023.

\bibitem[Saharia et~al.(2022)Saharia, Ho, Chan, Salimans, Fleet, and Norouzi]{sr3}
C. Saharia, J. Ho, W. Chan, T. Salimans, D.~J. Fleet, and M. Norouzi.
\newblock Image super-resolution via iterative refinement.
\newblock \emph{\PAMI}, PP, 2022.

\bibitem[Shen and Shen(2023)]{psrflow}
J. Shen and H.-W. Shen.
\newblock Psrflow: Probabilistic super resolution with flow-based models for scientific data.
\newblock \emph{IEEE Transactions on Visualization and Computer Graphics}, 2023.

\bibitem[Shi et~al.(2016)Shi, Caballero, Huszar, Totz, Aitken, Bishop, Rueckert, and Wang]{espcn}
W. Shi, J. Caballero, F. Huszar, J. Totz, A.~P. Aitken, R. Bishop, D. Rueckert, and Z. Wang.
\newblock Real-time single image and video super-resolution using an efficient sub-pixel convolutional neural network.
\newblock In \emph{\CVPR}, pages 1874--1883, 2016.

\bibitem[Sohl-Dickstein et~al.(2015)Sohl-Dickstein, Weiss, Maheswaranathan, and Ganguli]{sohl2015deep}
J. Sohl-Dickstein, E. Weiss, N. Maheswaranathan, and S. Ganguli.
\newblock Deep unsupervised learning using nonequilibrium thermodynamics.
\newblock In \emph{\ICML}, pages 2256--2265, 2015.

\bibitem[Song and Ermon(2019)]{song2019generative}
Y. Song and S. Ermon.
\newblock Generative modeling by estimating gradients of the data distribution.
\newblock \emph{\NIPS}, 32, 2019.

\bibitem[Song and Ermon(2020)]{song2020improved}
Y. Song and S. Ermon.
\newblock Improved techniques for training score-based generative models.
\newblock \emph{\NIPS}, 33:\penalty0 12438--12448, 2020.

\bibitem[Song et~al.(2020)Song, Sohl-Dickstein, Kingma, Kumar, Ermon, and Poole]{song2020score}
Y. Song, J. Sohl-Dickstein, D.~P. Kingma, A. Kumar, S. Ermon, and B. Poole.
\newblock Score-based generative modeling through stochastic differential equations.
\newblock \emph{\ICLR}, 2020.

\bibitem[Sung et~al.(2022)Sung, Shim, Kim, Lee, and Kim]{fsncsr}
K.-U. Sung, D. Shim, K.-W. Kim, J.-Y. Lee, and Y. Kim.
\newblock Fs-ncsr: Increasing diversity of the super-resolution space via frequency separation and noise-conditioned normalizing flow.
\newblock In \emph{\CVPRW}, pages 967--976. IEEE, 2022.

\bibitem[Timofte et~al.(2017)Timofte, Agustsson, Gool, Yang, and Zhang]{flickr2k}
R. Timofte, E. Agustsson, L.~Van Gool, M.-H. Yang, and L. Zhang.
\newblock {NTIRE} 2017 challenge on single image super-resolution: Methods and results.
\newblock In \emph{\CVPRW{}}, pages 1110--1121, 2017.

\bibitem[Tomczak and Welling(2018)]{tomczak2018vae}
J. Tomczak and M. Welling.
\newblock Vae with a vampprior.
\newblock In \emph{International Conference on Artificial Intelligence and Statistics}, pages 1214--1223, 2018.

\bibitem[Wang et~al.(2023)Wang, Chen, Ni, Liu, and Liu]{omnisr}
H. Wang, X. Chen, B. Ni, Y. Liu, and J. Liu.
\newblock Omni aggregation networks for lightweight image super-resolution.
\newblock In \emph{\CVPR}, pages 22378--22387, 2023.

\bibitem[Wang et~al.(2018{\natexlab{a}})Wang, Yu, Dong, and Loy]{sftgan}
X. Wang, K. Yu, C. Dong, and C.~C. Loy.
\newblock Recovering realistic texture in image super-resolution by deep spatial feature transform.
\newblock In \emph{\CVPR}, pages 606--615, 2018{\natexlab{a}}.

\bibitem[Wang et~al.(2018{\natexlab{b}})Wang, Yu, Wu, Gu, Liu, Dong, Qiao, and Loy]{esrgan}
X. Wang, K. Yu, S. Wu, J. Gu, Y. Liu, C. Dong, Y. Qiao, and C.~C. Loy.
\newblock Esrgan: Enhanced super-resolution generative adversarial networks.
\newblock In \emph{\ECCVW}, pages 63--79, 2018{\natexlab{b}}.

\bibitem[Wu et~al.(2022)Wu, Ni, Wang, and Zhang]{blindsrsnf}
H. Wu, N. Ni, S. Wang, and L. Zhang.
\newblock Blind super-resolution for remote sensing images via conditional stochastic normalizing flows.
\newblock \emph{arXiv preprint arXiv:2210.07751}, 2022.

\bibitem[Xu et~al.(2021)Xu, Wang, and Shi]{ultrasr}
X. Xu, Z. Wang, and H. Shi.
\newblock Ultrasr: Spatial encoding is a missing key for implicit image function-based arbitrary-scale super-resolution.
\newblock \emph{CoRR}, abs/2103.12716, 2021.

\bibitem[Yang et~al.(2021)Yang, Shen, Yue, and Li]{itsrn}
J. Yang, S. Shen, H. Yue, and K. Li.
\newblock Implicit transformer network for screen content image continuous super-resolution.
\newblock In \emph{\NIPS}, pages 13304--13315, 2021.

\bibitem[Yao et~al.(2023)Yao, Tsao, Lo, Tseng, Chang, and Lee]{linf}
J.-E. Yao, L.-Y. Tsao, Y.-C. Lo, R. Tseng, C.-C. Chang, and C.-Y. Lee.
\newblock Local implicit normalizing flow for arbitrary-scale image super-resolution.
\newblock In \emph{\CVPR{}}, pages 1776--1785, 2023.

\bibitem[Zeyde et~al.(2010)Zeyde, Elad, and Protter]{set14}
R. Zeyde, M. Elad, and M. Protter.
\newblock On single image scale-up using sparse-representations.
\newblock In \emph{Curves and Surfaces}, pages 711--730, 2010.

\bibitem[Zhang et~al.(2018{\natexlab{a}})Zhang, Isola, Efros, Shechtman, and Wang]{lpips}
R. Zhang, P. Isola, A.~A. Efros, E. Shechtman, and O. Wang.
\newblock The unreasonable effectiveness of deep features as a perceptual metric.
\newblock In \emph{\CVPR{}}, pages 586--595, 2018{\natexlab{a}}.

\bibitem[Zhang et~al.(2019)Zhang, Liu, Dong, and Qiao]{ranksrgan}
W. Zhang, Y. Liu, C. Dong, and Y. Qiao.
\newblock Ranksrgan: Generative adversarial networks with ranker for image super-resolution.
\newblock In \emph{\ICCV}, pages 3096--3105, 2019.

\bibitem[Zhang et~al.(2018{\natexlab{b}})Zhang, Li, Li, Wang, Zhong, and Fu]{rcan}
Y. Zhang, K. Li, K. Li, L. Wang, B. Zhong, and Y. Fu.
\newblock Image super-resolution using very deep residual channel attention networks.
\newblock In \emph{\ECCV}, pages 286--301, 2018{\natexlab{b}}.

\bibitem[Zhang et~al.(2018{\natexlab{c}})Zhang, Tian, Kong, Zhong, and Fu]{rdn}
Y. Zhang, Y. Tian, Y. Kong, B. Zhong, and Y. Fu.
\newblock Residual dense network for image super-resolution.
\newblock In \emph{\CVPR}, pages 2472--2481, 2018{\natexlab{c}}.

\bibitem[Zhou et~al.(2023)Zhou, Li, Guo, Bai, Cheng, and Hou]{srformer}
Y. Zhou, Z. Li, C.-L. Guo, S. Bai, M.-M. Cheng, and Q. Hou.
\newblock Srformer: Permuted self-attention for single image super-resolution.
\newblock \emph{\ICCV}, 2023.

\end{thebibliography}
}


\end{document}


\maketitle


\section{Overview}
\label{sec::supp_overview}
This supplementary material first provides an expanded notation table used in this work. Following this, more detailed information is presented for enhanced clarity, including additional experimental results and implementation details. Lastly, more qualitative comparisons are shown to demonstrate the capabilities of the proposed framework.


\section{Notation Table}
\label{sec::supp_notation}
\begin{table*}[t]
\renewcommand{\arraystretch}{1.3}
\newcommand{\mytoprule}{\toprule[1.2pt]}
\centering
\footnotesize
\setlength{\tabcolsep}{1.3em}
\caption{The notations used in this study and their corresponding definitions.}
\resizebox{2\columnwidth}{!}{%
    \begin{tabular}{c|l}
        \mytoprule
        Notation & Definition  \\
        \mytoprule
        $N$ & The size of a dataset. \\
        $x$ & Low-resolution image. \\
        $x^{up}$ & Bilinear upsampled low-resolution image. \\
        $y$ & High-resolution image. \\
        $\hat{y}$ & High-resolution image generated by the flow model. \\
        $m$ & The residual map between $y$ and $x_{up}$. \\
        $z$ & A latent code in a standard normal distribution (i.e., $z \in \mathcal{N}(0, I)$). \\
        $f_{\theta}$ & Pre-trained flow-based super-resolution model~\cite{srflow, linf}, which is parameterized by $\theta$. \\
        $\tau$ & The sampling temperature (i.e., the standard deviation of a Gaussian distribution). \\
        $s$ & The scaling factor. \\
        $c$ & The coordinate of a patch~\cite{linf}. \\
        $n$ & The size of a patch~\cite{linf}. \\
        \hline
        $G$ & The proposed latent module. \\
        $\hat{z}$ & The learned prior generated by $G$. \\
        $z^*$ & The latent code corresponding to the ground truth image $y$, transformed by $f_\theta$. \\
        $\mathcal{L}_{latent}$ & The L1 loss calculated between $z^*$ and $\hat{z}$. \\
        $\mathcal{L}_{percep}$ & The perceptual loss~\cite{vggloss} calculated between $y$ and $\hat{y}$. \\
        $\mathcal{L}_{total}$ & The final objective function. \\
        $\Psi_{per}$ & A pre-trained VGG19~\cite{vgg} network. \\
        \mytoprule
    \end{tabular}
}
\label{tab:supp:notation}
\end{table*}
Table~\ref{tab:supp:notation} presents the notations used in the main text and their corresponding definitions.

\section{Detailed Analyses}
\label{sec::supp_detail}
This section analyzes the design flexibility of our latent module in arbitrary-scale super-resolution (SR) tasks and demonstrates the visual effects of employing different objective functions.

\paragraph{Design flexibility in Arbitrary-scale SR tasks.}
{
\begin{table*}[t]
\caption{The arbitrary-scale SR results on SR benchmark datasets. The names in the parentheses (\eg, UNet) refer to the architecture of our latent generator. ``\textit{In-scales}'' and ``\textit{OOD-scales}'' refer to in- and out-of-training-distribution scales. LPIPS~\cite{lpips} scores are reported (lower is better), with the best and second-best highlighted in \textbf{\red{red}} and \blue{blue}, respectively.}
\newcommand{\mytoprule}{\toprule[1.2pt]}
\renewcommand{\arraystretch}{1.6}
\centering
\setlength{\tabcolsep}{1.0em}
\resizebox{2.1\columnwidth}{!}{%
    \Huge
    \begin{tabular}{l|ccc|cc|ccc|cc|ccc|cc|ccc|cc}
    \toprule
     & \multicolumn{5}{c|}{Set5~\cite{set5}} & \multicolumn{5}{c|}{Set14~\cite{set14}} & \multicolumn{5}{c|}{B100~\cite{b100}} & \multicolumn{5}{c}{Urban100~\cite{urban100}} \\ \cline{2-21}
     
     Method & 
     \multicolumn{3}{c|}{In-scales} & \multicolumn{2}{c|}{OOD-scales} & \multicolumn{3}{c|}{In-scales} & \multicolumn{2}{c|}{OOD-scales} &
     \multicolumn{3}{c|}{In-scales} & \multicolumn{2}{c|}{OOD-scales} & \multicolumn{3}{c|}{In-scales} & \multicolumn{2}{c}{OOD-scales} \\
     
     & $\times$2 & $\times$3 & $\times$4 & $\times$6 & $\times$8
     & $\times$2 & $\times$3 & $\times$4 & $\times$6 & $\times$8
     & $\times$2 & $\times$3 & $\times$4 & $\times$6 & $\times$8
     & $\times$2 & $\times$3 & $\times$4 & $\times$6 & $\times$8 \\
     
     \hline \hline
     
     EDSR-baseline-MetaSR~\cite{metasr} & 0.057 & 0.125 & 0.175 & 0.253 & 0.326 & 0.094 & 0.207 & 0.286 & 0.395 & 0.460 & 0.147 & 0.285 & 0.376 & 0.492 & 0.565 & 0.065 & 0.157 & 0.233 & 0.352 & 0.446 \\
     
     EDSR-baseline-LIIF~\cite{liif} & 0.056 & 0.124 & 0.173 & 0.248 & 0.307 & 0.093 & 0.205 & 0.284 & 0.390 & 0.449 & 0.147 & 0.282 & 0.372 & 0.486 & 0.556 & 0.064 & 0.155 & 0.228 & 0.338 & 0.422 \\
     
     EDSR-baseline-LTE~\cite{lte} & 0.056 & 0.123 & 0.174 & 0.257 & 0.326 & 0.092 & 0.203 & 0.283 & 0.396 & 0.463 & 0.146 & 0.280 & 0.371 & 0.495 & 0.570 & 0.063 & 0.152 & 0.224 & 0.345 & 0.436 \\
     
     EDSR-baseline-LINF~\cite{linf} ($\tau$ = $\tau_0$) & 0.035 & 0.067 & 0.088 & 0.158 & \textcolor{blue}{0.249} & 0.064 & 0.115 & 0.163 & 0.275 & \textcolor{blue}{0.375} & 0.108 & 0.172 & 0.207 & 0.319 & 0.451 & 0.050 & 0.110 & 0.158 & 0.273 & 0.386 \\
     
     \textbf{EDSR-baseline-Ours (EDSR-baseline)} & \textbf{\textcolor{red}{0.026}} & \textcolor{blue}{0.051} & \textbf{\textcolor{red}{0.072}} & 0.157 & 0.276 & \textbf{\textcolor{red}{0.052}} & \textcolor{blue}{0.097} & 0.147 & 0.269 & 0.390 & \textbf{\textcolor{red}{0.083}} & \textcolor{blue}{0.132} & \textcolor{blue}{0.177} & 0.304 & 0.440 & \textbf{\textcolor{red}{0.043}} & \textbf{\textcolor{red}{0.097}} & \textcolor{blue}{0.144} & 0.260 & 0.388 \\
     
     \textbf{EDSR-baseline-Ours (UNet)} & \textbf{\textcolor{red}{0.026}} & \textbf{\textcolor{red}{0.047}} & 0.074 & \textbf{\textcolor{red}{0.145}} & \textbf{\textcolor{red}{0.243}} & \textcolor{blue}{0.054} & \textbf{\textcolor{red}{0.094}} & \textcolor{blue}{0.144} & \textcolor{blue}{0.253} & \textbf{\textcolor{red}{0.364}} & \textcolor{blue}{0.084} & \textbf{\textcolor{red}{0.127}} & \textcolor{blue}{0.177} & \textcolor{blue}{0.289} & \textcolor{blue}{0.425} & 0.440 & 0.098 & 0.146 & \textcolor{blue}{0.253} & \textbf{\textcolor{red}{0.377}} \\
     
     \textbf{EDSR-baseline-Ours (Swin-T)} & 0.029 & 0.053 & \textcolor{blue}{0.073} & \textcolor{blue}{0.149} & 0.277 & 0.055 & 0.100 & \textbf{\textcolor{red}{0.141}} & \textbf{\textcolor{red}{0.252}} & 0.379 & 0.090 & 0.135 & \textbf{\textcolor{red}{0.176}} & \textbf{\textcolor{red}{0.287}} & \textbf{\textcolor{red}{0.422}} & \textbf{\textcolor{red}{0.043}} & \textbf{\textcolor{red}{0.097}} & \textbf{\textcolor{red}{0.141}} & \textbf{\textcolor{red}{0.251}} & \textcolor{blue}{0.382} \\
     
     \hline \hline
     
     RRDB-LINF~\cite{linf} ($\tau$ = $\tau_0$) & 0.034 & 0.064 & 0.084 & 0.147 & \textcolor{blue}{0.247} & 0.059 & 0.110 & 0.146 & 0.252 & 0.359 & 0.097 & 0.152 & 0.194 & 0.306 & 0.444 & 0.040 & 0.093 & 0.137 & 0.239 & 0.354 \\
     
     \textbf{RRDB-Ours (EDSR-baseline)} & \textcolor{blue}{0.025} & \textcolor{blue}{0.044} & \textcolor{blue}{0.066} & 0.136 & 0.249 & \textcolor{blue}{0.046} & \textcolor{blue}{0.087} & 0.129 & 0.230 & 0.329 & \textcolor{blue}{0.069} & 0.118 & 0.165 & 0.272 & 0.385 & 0.035 & 0.082 & \textcolor{blue}{0.126} & 0.230 & 0.352 \\
     
     \textbf{RRDB-Ours (Unet)} & \textbf{\textcolor{red}{0.023}} & \textbf{\textcolor{red}{0.042}} & \textcolor{blue}{0.066} & \textcolor{blue}{0.131} & \textbf{\textcolor{red}{0.234}} & \textbf{\textcolor{red}{0.043}} & \textcolor{blue}{0.087} & \textcolor{blue}{0.124} & \textbf{\textcolor{red}{0.221}} & \textbf{\textcolor{red}{0.322}} & \textbf{\textcolor{red}{0.061}} & \textbf{\textcolor{red}{0.113}} & \textcolor{blue}{0.163} & \textbf{\textcolor{red}{0.264}} & \textbf{\textcolor{red}{0.378}} & \textbf{\textcolor{red}{0.033}} & \textcolor{blue}{0.081} & \textcolor{blue}{0.126} & \textbf{\textcolor{red}{0.219}} & \textbf{\textcolor{red}{0.331}} \\
     
     \textbf{RRDB-Ours (Swin-T)} & \textcolor{blue}{0.025} & 0.045 & \textbf{\textcolor{red}{0.063}} & \textbf{\textcolor{red}{0.129}} & \textcolor{blue}{0.247} & \textcolor{blue}{0.046} & \textbf{\textcolor{red}{0.086}} & \textbf{\textcolor{red}{0.123}} & \textcolor{blue}{0.223} & \textcolor{blue}{0.328} & 0.071 & \textcolor{blue}{0.115} & \textbf{\textcolor{red}{0.162}} & \textcolor{blue}{0.267} & \textcolor{blue}{0.383} & \textcolor{blue}{0.034} & \textbf{\textcolor{red}{0.080}} & \textbf{\textcolor{red}{0.123}} & \textcolor{blue}{0.221} & \textcolor{blue}{0.343} \\
     
    \bottomrule
    \end{tabular}
}
\label{tab:supp:arbit_sr}
\vspace{-5pt}
\end{table*}
}
In the main paper, we present the arbitrary-scale SR results of our proposed framework, which employs UNet~\cite{unet} as the latent generator. To further validate the design flexibility of our latent module, we provide results in arbitrary-scale SR tasks using EDSR-baseline~\cite{edsr} and Swin Transformer~\cite{swin} (Swin-T) as alternative latent generators. This analysis includes two variants of our model: EDSR-baseline-LINF-LP and RRDB-LINF-LP, which facilitates a comprehensive comparison. For clarity, we denote them as EDSR-baseline-Ours and RRDB-Ours, respectively. The results in Table~\ref{tab:supp:arbit_sr} demonstrate that our framework achieves promising results with all these backbones as the latent generator, proving the flexibility of our framework in arbitrary-scale SR tasks.

\paragraph{Visual Effects with Different Objective Functions.}
\begin{figure}[t]
  \centering
  \includegraphics[width=0.75\linewidth]{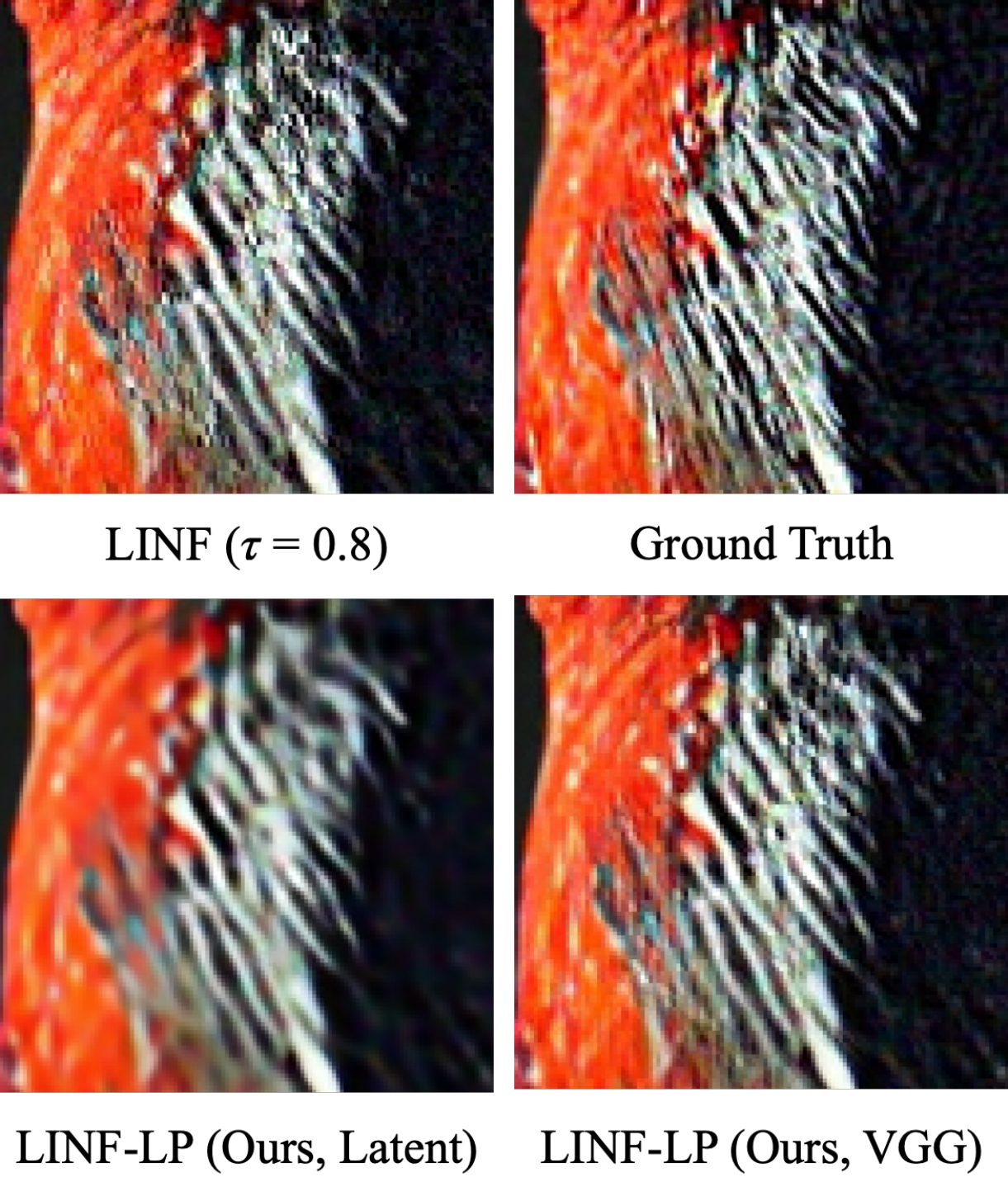}
  \caption{
    A qualitative comparison between the effects of employing different objective functions. ``Latent" represents the exclusive use of latent space loss, while ``VGG" denotes the sole employment of VGG perceptual loss~\cite{vggloss}.
  }
  \label{fig:supp_loss}
\end{figure}
Figure~\ref{fig:supp_loss} demonstrates the qualitative results obtained by training our latent module using different objective functions. As shown in this figure, our framework generates sharper content when employing the perceptual loss, in contrast to the smooth but blurry results with the latent space loss. In addition, our framework effectively mitigates the grid artifacts presented in image produced by an LINF~\cite{linf} model.

\section{Implementation Details}
\label{sec::supp_implement}
In this section, we illustrate the derivation of the ``best temperature map" and additional details of SRFlow-LP.

\paragraph{Best Temperature Map.}
We adapt the derivation of the ``Optimal Objective Estimation" in SROOE~\cite{srooe} to generate our ``best temperature map".  Specifically, for each image, we generate a total of 21 outcomes using an LINF model, with sampling temperatures ranging from 0 to 1 at intervals of 0.05. Then, for every pixel in each image, we compute the LPIPS~\cite{lpips} values and select the temperature that yields the optimal LPIPS from these 21 images. Once the optimal temperature for every pixel is determined, we assemble these selections to form a ``best temperature map". Note that in \cite{srooe}, they search a $t$ value as a conditional input to their GAN model, while in our approach, the temperature $\tau$ represents the standard deviation of a Gaussian distribution.

\paragraph{SRFlow-LP Implementation Details.}
We found that employing the latent space loss as a regularization term for training SRFlow-LP effectively prevents exploding inverses. This effectiveness stems from the observation that models trained with the latent space loss tend to produce latent codes that fall within the training distribution, therefore avoiding subsequent exploding inverses. To further stabilize the inference process of SRFlow-LP, the initial prior is normalized before being processed by the latent module. In addition, we skip the iteration which encounters an exploding inverse during training. These techniques allow SRFlow-LP to be more stable during both training and inference without modifying the architecture or inference pipeline of the proposed framework.

\section{Additional Qualitative Results}
\label{sec::supp_qual}
\subsection{Qualitative Results of SRFlow-LP}
\begin{figure*}[t]
  \centering
  \includegraphics[width=0.7\linewidth]{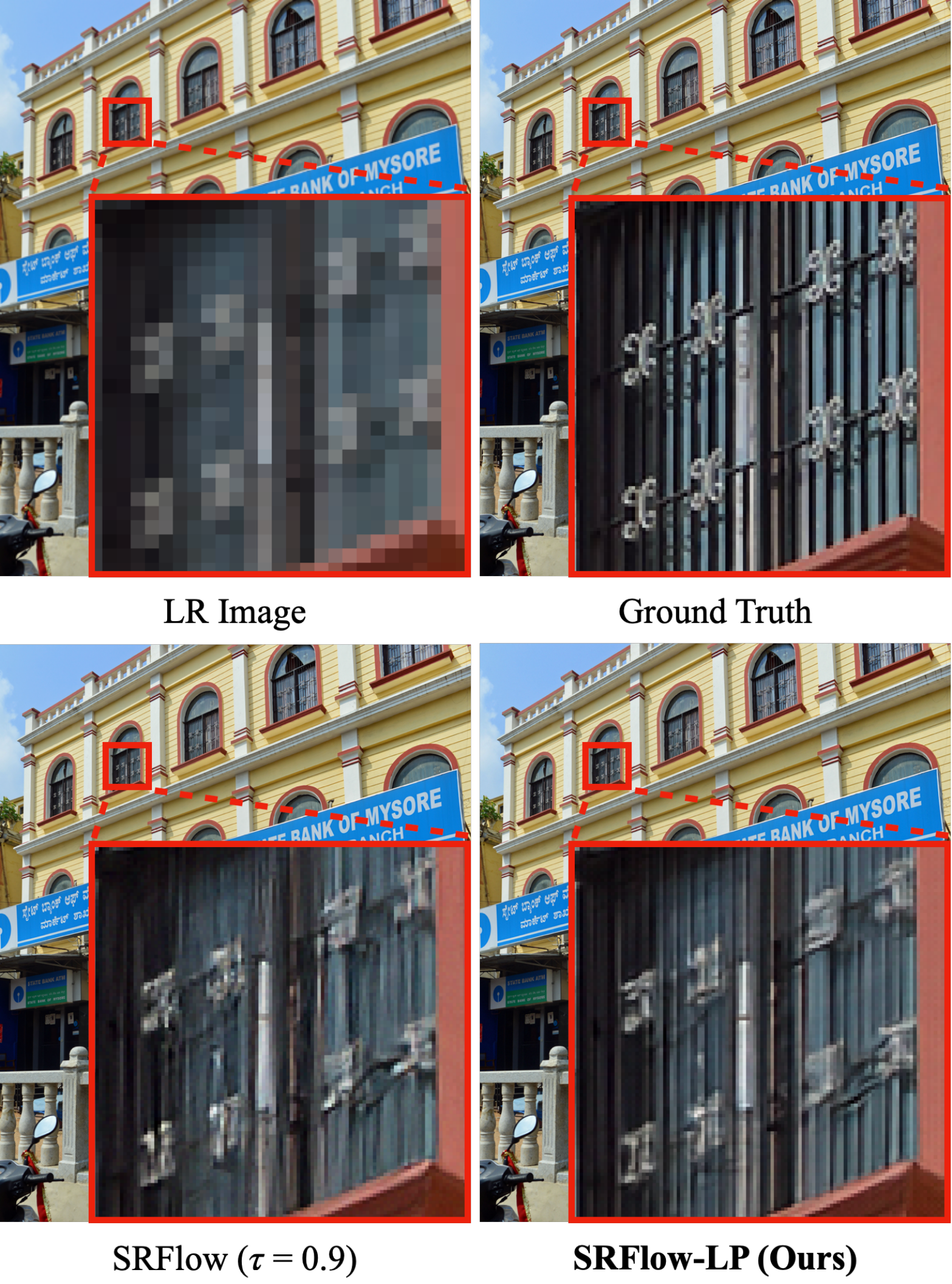}
  \caption{
    A qualitative comparison between the 4$\times$ SR results of SRFlow~\cite{srflow} and our SRFlow-LP. 
  }
  \label{fig:supp_srflow_window}
\end{figure*}

\begin{figure*}[t]
  \centering
  \includegraphics[width=0.8\linewidth]{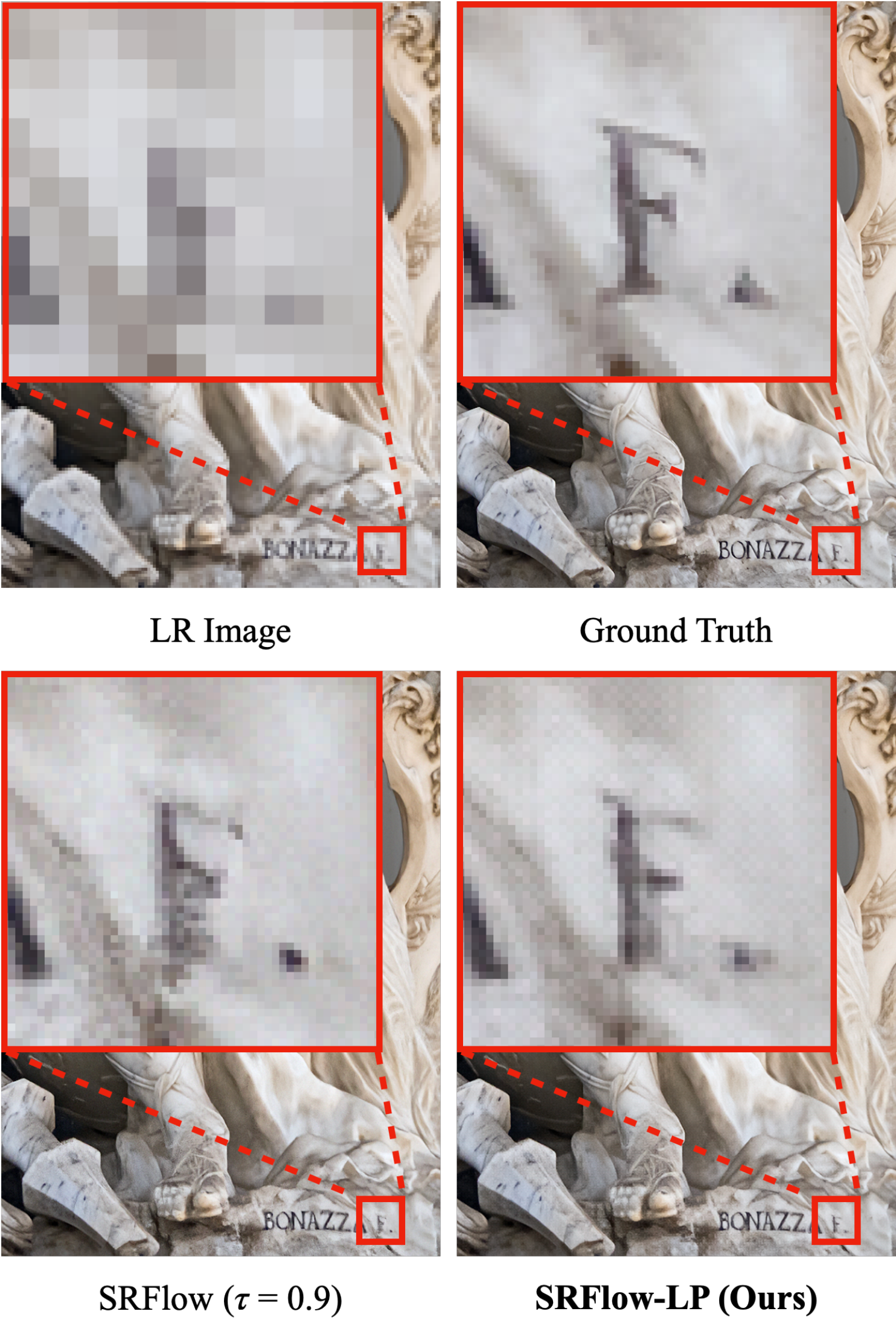}
  \caption{
    A qualitative comparison between the 4$\times$ SR results of SRFlow~\cite{srflow} and our SRFlow-LP. 
  }
  \label{fig:supp_srflow_f}
\end{figure*}

\begin{figure*}[t]
  \centering
  \includegraphics[width=0.8\linewidth]{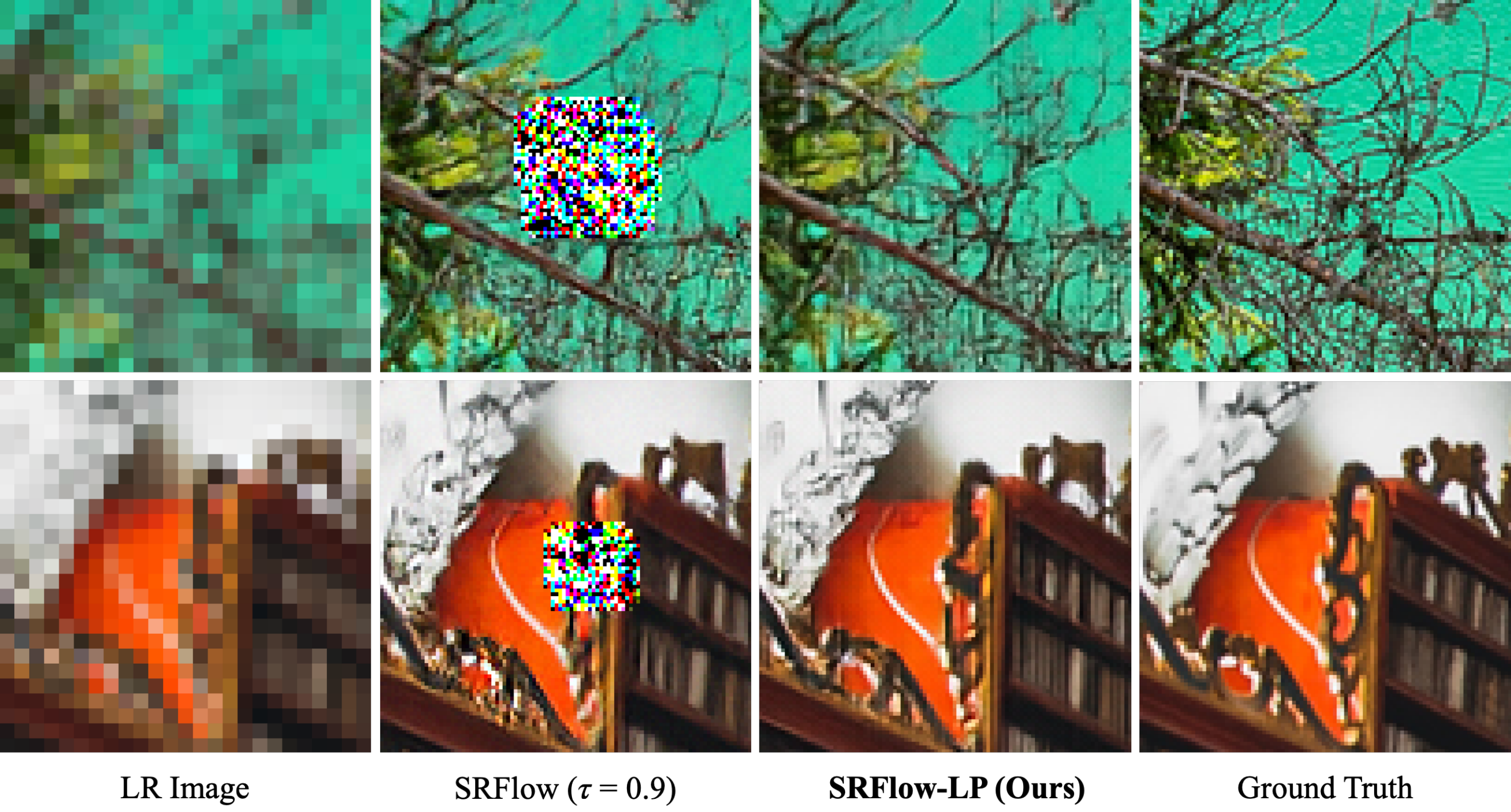}
  \caption{
    Our framework SRFlow-LP effectively prevents SRFlow~\cite{srflow} from encountering exploding inverses. 
  }
  \label{fig:supp_srflow_inf}
\end{figure*}

Figs.~\ref{fig:supp_srflow_window} and~\ref{fig:supp_srflow_f} demonstrate our SRFlow-LP generates images with sharper details than the original SRFlow~\cite{srflow}. Fig.~\ref{fig:supp_srflow_inf} also presents that SRFlow-LP effectively prevents SRFlow from encountering exploding inverses that display noisy patches within images.

\subsection{Qualitative Results of LINF-LP}
\begin{figure*}[t]
  \centering
  \includegraphics[width=0.9\linewidth]{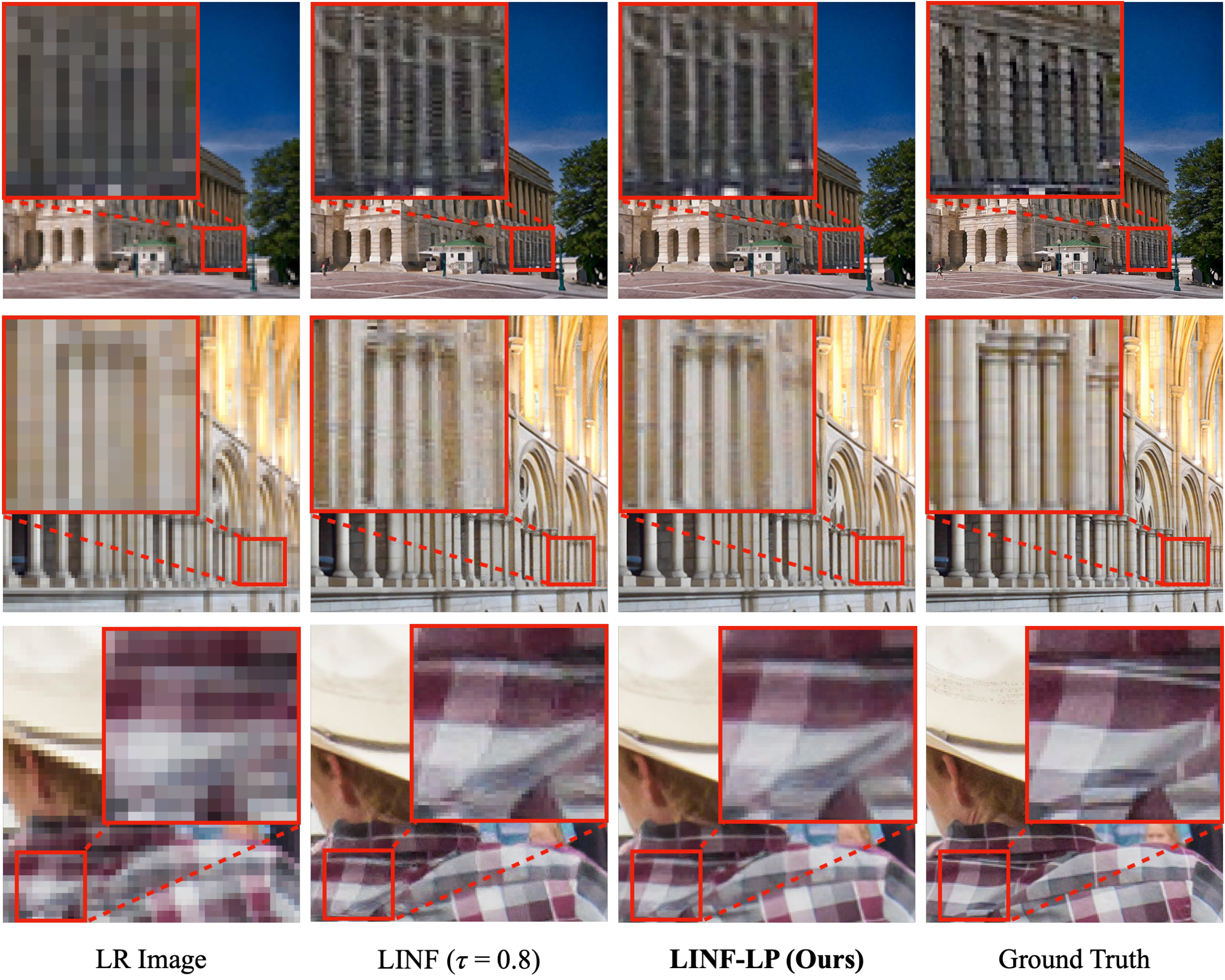}
  \caption{
    A qualitative comparison between images generated by LINF~\cite{linf} and LINF-LP.
  }
  \label{fig:supp_linf1}
\end{figure*}

Fig.~\ref{fig:supp_linf1} illustrates that our LINF-LP mitigates the grid artifacts in images generated by LINF~\cite{linf}, especially in areas with thin, repetitive linear structures. 

{
    \small
    \bibliographystyle{ieeenat_fullname}
    \bibliography{main}
}
